\documentclass[conference]{IEEEtran}
\usepackage{times}

\usepackage[numbers]{natbib}
\usepackage{multicol}
\usepackage[bookmarks=true]{hyperref}
\usepackage{graphics} 
\usepackage{graphicx}
\usepackage{subcaption} 
\usepackage{epsfig} 
\usepackage{times} 
\usepackage{amsmath} 
\usepackage{amssymb}  
\usepackage{algorithm}
\usepackage{algorithmic}
\usepackage{bm}
\usepackage{hyperref}
\usepackage{multirow}

\DeclareMathOperator*{\argmin}{argmin}
\newcommand{\etal}{\textit{et al}.}

\begin{document}

\title{Batch and Incremental Kinodynamic Motion Planning using Dynamic Factor Graphs}


\author{\authorblockN{Mandy Xie and Frank Dellaert}
\authorblockA{School of Interactive Computing, 
Georgia Institute of Technology\\
Atlanta, Georgia 30332--0250, 
Email: \{manxie,dellaert\}@gatech.edu}
}


%

\maketitle

\begin{abstract}
This paper presents a kinodynamic motion planner that is able to produce energy efficient motions by taking the full robot dynamics into account, and making use of gravity, inertia, and momentum to reduce the effort. Given a specific goal state for the robot, we use factor graphs and numerical optimization to solve for an optimal trajectory, which meets not only the requirements of collision avoidance, but also all kinematic and dynamic constraints, such as velocity, acceleration and torque limits. By exploiting the sparsity in factor graphs, we can solve a kinodynamic motion planning problem efficiently, on par with existing optimal control methods, and use incremental elimination techniques to achieve an order of magnitude faster replanning.
\end{abstract}

\IEEEpeerreviewmaketitle

\section{Introduction}

Kinodynamic motion planning is an important and active research area in robotics \cite{Donald93jacm_KDMP, Donald95algorithmica_kino, Lavalle01ijrr_kinoRRT, Hsu02ijrr_kinoEST, Lavalle06book_rrt, Perez12icra_lqrRRT, Schmerling19eor_KDMP}. 
Purely kinematic motion planning focuses on finding a trajectory through a robot's configuration space that satisfies multiple criteria such as collision-avoidance, obeying joint limits, and honoring smoothness constraints. 
However, to produce energy-efficient motions and satisfy dynamic constraints such as force and torque limits, we need \textit{kinodynamic} motion planners that take the full robot dynamics into account. In addition, many tasks involving large payloads, fast motions, or both, require inherently dynamic behaviors that cannot be achieved by reasoning about kinematics alone.

\begin{figure}
	\centering
	\includegraphics[scale=1]{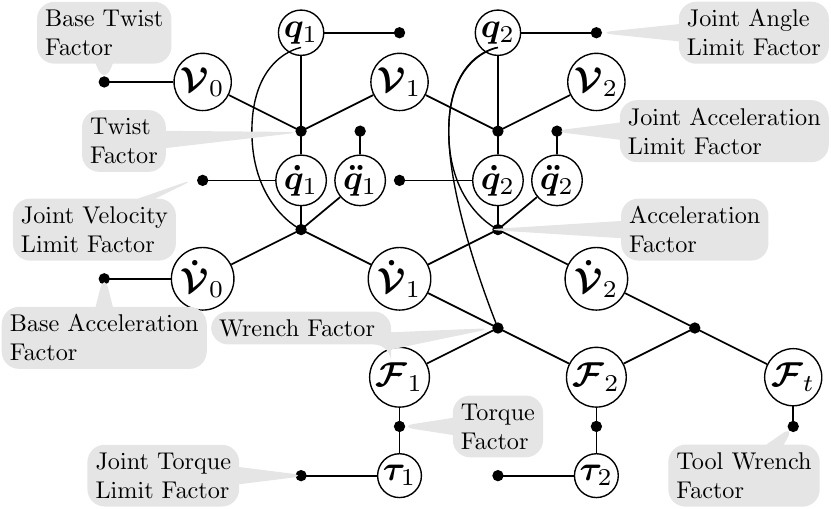}
	\caption{The dynamic factor graph (DFG) for a RR manipulator, where black dots represent factors, and circles represent variables. Figure adapted from~\cite{Xie19arxiv_dynamicsFactorGraph}.}
	\label{fig:dfg}
\end{figure}

\begin{figure}[!ht]
	\centering
	\includegraphics[scale=1]{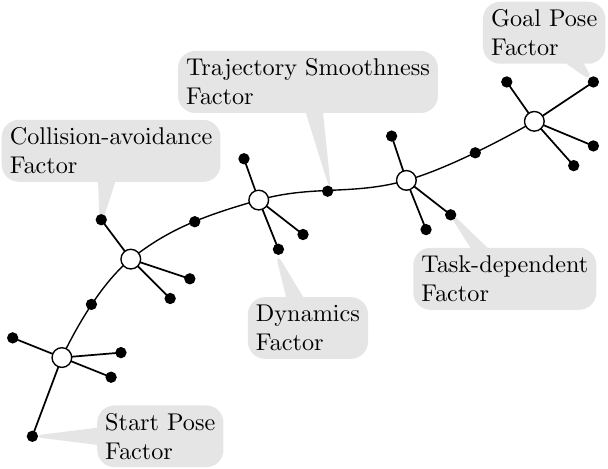}
	\caption{Kinodynamic motion planning factor graph. To simplify the notation, we use a dynamics factor to represent all the dynamic constraints at each time slice in the factor graph.}
	\label{fig:dfgp}
\end{figure}

In this paper we consider the kinodynamic motion planning as an optimal control (OC) problem, using factor graphs as a convenient and intuitive representation that also provides distinct computational advantages.
In using OC-techniques we are inspired by the recent work by Zhao \etal~\cite{Zhao18icarcv_optControlMP} for manipulators, and many other OC-style approaches from the dynamic walking literature~\cite{Khoury13icra_optControlMP,Lengagne13ijrr_optKMP}. 
In using a factor graph representation we follow the lead of GPMP2~\cite{Mukadam18ijrr_gpmp2}, which solves the \textit{kinematic} motion planning problem efficiently using a sparse nonlinear solver~\cite{Dellaert12report_gtsam, Dellaert17fnt_fg}, and performs fast replanning by taking advantage of an incremental solver~\cite{Kaess12ijrr}.

To incorporate dynamics, we represent kinodynamic constraints with a \textit{dynamic factor graph} (DFG) as introduced by Xie and Dellaert~\cite{Xie19arxiv_dynamicsFactorGraph}.
Fig.~\ref{fig:dfg} shows the DFG of a RR manipulator, where the equations of motion are represented as factors, and variables involved in the equations are represented as nodes in the graph.
We then construct an optimal control problem by adding factors representing all the usual trajectory optimization objectives, as well as factors enforcing the dynamics between successive time-slices, as shown in Figure~\ref{fig:dfgp}.

By combining optimal control and factor graphs we obtain an intuitive, fast kinodynamic motion planner that is on par with the fastest OC methods, and adds efficient, incremental kinodynamic replanning.
The use of GTSAM~\cite{Dellaert12report_gtsam, Dellaert17fnt_fg} as an optimizer ensures that we are exploiting the benefits of both sparsity and automatic differentiation, which was shown to be crucial to achieving state of the art performance by Zhao \etal~\cite{Zhao18icarcv_optControlMP}. 
Fast re-planning is useful in many contexts, e.g., model-predictive control (MPC).
And, as far as we know, no existing kinodynamic motion planning methods can perform fast replanning beyond warm-starting as used in MPC.

\section{Related Work}

Kinodynamic motion planning has a rich history. The term was coined in the papers by Donald~\etal~\cite{Donald93jacm_KDMP,Donald95algorithmica_kino}, and the well known Rapidly-exploring Random Trees (RRTs) were in fact developed with kinodynamic constraints in mind from the very start~\cite{LaValle99icra_kinodynamic_rrt, Lavalle01ijrr_kinoRRT}, as were the Expansive Space Tree algorithms by Hsu~\etal~\cite{Hsu97icra_est, Hsu02ijrr_kinoEST}.
In many cases, however, sampling-based methods such as RRTs and Probabilistic Road Maps (PRM)~\cite{Kavraki96tra_prm, Amato96icra_prm} are used for purely kinematic motion planning. 
Satisfying both kinematic \textit{and} dynamic constraints simultaneously makes the kinodynamic motion planning problem more challenging. 
Instead of planning in the configuration space, we now need to plan in the state space, whose dimension is twice that of the configuration space.

More importantly, kinodynamic sampling-based methods need both good \textit{distance metrics} and efficient \textit{steering methods}~\cite{Choset05book_robot,Allen19ras_realtimeKDMP}. A distance metric is needed in RRT-style methods to choose which part of the tree to expand, and simple metrics like the Euclidean distance do not incorporate any knowledge of the system dynamics~\cite{Perez12icra_lqrRRT}.
The other issue is the lack of computationally efficient steering methods to connect two states
to produce a segment between them 
in a way that satisfies all differential constraints~\cite{Kunz14iros_kinodynamic}.
LQR-RRT*~\cite{Perez12icra_lqrRRT} addresses both of these issues by linearizing the system dynamics and using the LQR cost function as the distance metric, 
similar to the earlier work by Glassman and Tedrake~\cite{Glassman10icra_lqrTree}.
Webb \etal~\cite{Webb13icra_kinodynamicRRT} improve on this by connecting any pair of states exactly via a fixed-final-state-free-final-time OC formulation for systems with controllable linear dynamics. Unfortunately, this still requires a linearized version of the dynamics system. More recently, the AO-RRT method proposed by Hauser \etal~\cite{Hauser16tor_kinoSample} and the SST and SST* methods presented by Li \etal~\cite{Li16ijrr_kinoSample} are able to solve kinodynamics motion planning problems for systems with a few degrees of freedom. 

Optimization-based motion planning methods have recently gained in popularity as they complement sampling-based methods~\cite{Zucker13ijrr_chomp, Kuindersma16ar_atlas, Schmerling19eor_KDMP}.
Hence, while sampling methods sometimes produce inconsistent and non-smooth solutions ~\cite{Pavlichenko17iros_stochasticOptimization}, they are often used to provide a feasible path, after which trajectory optimization is used to smooth out the trajectory. 
Examples of optimization-based methods include CHOMP~\cite{Zucker13ijrr_chomp}, STOMP~\cite{Kalakrishnan11icra_stomp}, TrajOpt~\cite{Schulman13rss_trajOpt}~\cite{Schulman14ijrr_trajOpt}, GPMP~\cite{Mukadam16icra_GPMP}, and GPMP2~\cite{Mukadam18ijrr_gpmp2}.
However, when applied to kinodynamic motion planning, 
the non-convex nature of the objective function, caused by dynamic constraints and constraints representing the obstacle-free region~\cite{Bergman19thesis_optControlMP}, makes these prone to converge to local minima unless properly initialized.
To deal with the difficulty of finding optimal solutions that satisfy all the constraints simultaneously,
Sintov \etal~\cite{Sintov19ar_optimization} generate random points within the allowable regions of the free parameters to find a set of feasible solutions, and use a gradient descent approach to refine the solution to a nearby local optimum. However, this approach is computationally demanding.

The kinodynamic motion planning problem can also be tackled with numerical optimal control methods~\cite{Bergman19thesis_optControlMP}, which enable easy incorporation of constraints and straightforward definition of objective functions. Based on optimal control techniques, Zhao \etal~\cite{Zhao18icarcv_optControlMP} present a motion planning approach, which uses pseudospectral method for trajectory discretization and interior point method with automatic differentiation for optimization. They solved the motion planning problem of a 6-axis robot with dynamic constraints, and it is shown to be much faster than existing works such as~\cite{Chettibi04ejmas_minimumTrajPlanning, Diehl06book_optControl}. A key element to achieving state of the art performance is the use of automatic differentiation over numerical or symbolic differentiation. Optimal control methods have also been applied to dynamic walking~\cite{Jelavic19iros_optControlWalking, Winkler18ral_gait} and humanoid robots motion planning~\cite{Khoury13icra_optControlMP,Lengagne13ijrr_optKMP}.

Finally, there exists an alternative way to deal with kinodynamic motion planning, which is to decouple the problem into a geometric part followed by re-timing. 
For instance, Hauser \etal~\cite{Hauser14ijrr_trajOptimization} use a dynamic interpolation method followed by a convex time-scaling optimization problem. Pham \etal~\cite{Pham17ijrr_optimzation} solve kinodynamic constraints with Admissible Velocity Propagation (AVP) based on Time-Optimal Path Parameterization (TOPP)~\cite{Bobrow85ijrr_topp}, which can be combined with sampling-based method such as RRT. However, this method may lead to sub-optimal solutions which sometimes may even be dynamically unfeasible~\cite{Bordalba17arxiv_constraintManifolds, Zhao18icarcv_optControlMP}.  

\section{Review: Motion Planning with Factor Graphs}

In this part, we review GPMP2~\cite{Mukadam18ijrr_gpmp2}, a purely kinematic motion planning method that optimizes for a collision-free \textit{and} smooth trajectory subject to a set of constraints. GPMP2 uses factor graphs to represent the corresponding objective functions, by making use of an incremental Bayes tree solver~\cite{Kaess12ijrr}, which supports fast replanning under changing conditions and/or updated goal states.

\subsection{Fixed Start and Goal Poses}
Since many applications demand a manipulator to move the end-effector from a start pose to a goal pose, both of the pose constraints are added to the motion planning problem. The cost of start and goal pose constraints can be expressed as a function of the state vectors $x_s$ and $x_g$:
\begin{align}
h_1(x_s) &= f(x_s) - p_s \label{eq:startpose}
\\
h_2(x_g) &= f(x_g) - p_g \label{eq:goalpose}
\end{align}
where $f()$ is the forward kinematics which maps any configuration to a workspace (this definition applies to the rest of the paper), $p_s$ is the desired start pose and $p_g$ is the desired goal pose of the end effector.

\subsection{Collision-avoidance}
To efficiently check collision for an arbitrarily shaped robot body, GPMP2 uses the method presented in CHOMP~\cite{Zucker13ijrr_chomp}, in which a robot's body is simplified to be a set of spheres, and the distance of the body to any point in the workspace is given by the distance to the center of the sphere minus its radius. The obstacles are represented by precomputed Signed Distance Field (SDF) matrices. The distance is positive if a point is outside of the obstacle, zero if it is on the surface, and negative when it lies inside the obstacle.

Hence, the obstacle cost function $h_3(x_i)$ is obtained by computing the signed distances for the robot in state $x_i$:
\begin{equation}\label{eq:obs}
h_3(x_i) = c_{sdf}(f(x_i))
\end{equation}
where $f()$ is the forward kinematics, and $c_{sdf}()$ is defined as the hinge loss:
\begin{equation}\label{eq:cz}
c_{sdf}(z)=\left\{
\begin{array}{ll}
-d(z) + \varepsilon\ &{if \ d(z) \leq \varepsilon}\\
0\ &{o/w}
\end{array}\right.
\end{equation}
where $d(z)$ is the signed distance from any point z in the workspace to the closest obstacle surface, which is obtained from a pre-computed SDF. Here we add a threshold $\varepsilon$ to prevent the robot from getting too close to obstacles.

\subsection{Task-dependent Kinematic Constraints}
In addition to the requirements of goal accomplishment, collision avoidance, and smoothness, the planner will sometimes have to satisfy user-specified constraints such as keeping the end-effector at a certain pose along the trajectory. For example, the cost of an end effector pose constraint can be expressed as a function of the state vector $x_i$:
\begin{equation}\label{eq:taskpose}
h_4(x_i) = f(x_i) - p_e
\end{equation}
where $f()$ is the forward kinematics, and $p_e$ is the desired end effector pose of the robot.

\subsection{Trajectory Smoothness}
Finally, \cite{Mukadam18ijrr_gpmp2} add a Gaussian process (GP) smoothness prior with cost function  
\begin{equation}\label{eq:smoothness}
h_5(x_{i-1}, x_i) = x_i - \Phi(t_i,t_{i-1})x_{i-1}
\end{equation}
and covariance matrix 
\begin{equation}\label{eq:sigma}
\Sigma_5 = \begin{bmatrix} \frac{1}{3} \Delta t_i^3 \mathbf{Q}_C &
\frac{1}{2} \Delta t_i^2 \mathbf{Q}_C \\ 
\frac{1}{2} \Delta t_i^2 \mathbf{Q}_C &
\Delta t_i \mathbf{Q}_C \end{bmatrix},
\end{equation}
where $Q_c$ is the power-spectral density matrix associated with the GP, and the state transition matrix 
\begin{equation}\label{eq:phi}
\Phi(t_i,t_{i-1}) = \begin{bmatrix} 1 & \Delta t_i\\ 0 & 1 \end{bmatrix}
\end{equation}
is associated with a constant velocity assumption between times $t_{i-1}$ and $t_i$, and $\Delta t_i = t_i - t_{i-1}.$

\subsection{Optimization with Factor Graphs}

\begin{figure}
	\centering
	\includegraphics[scale=1.2]{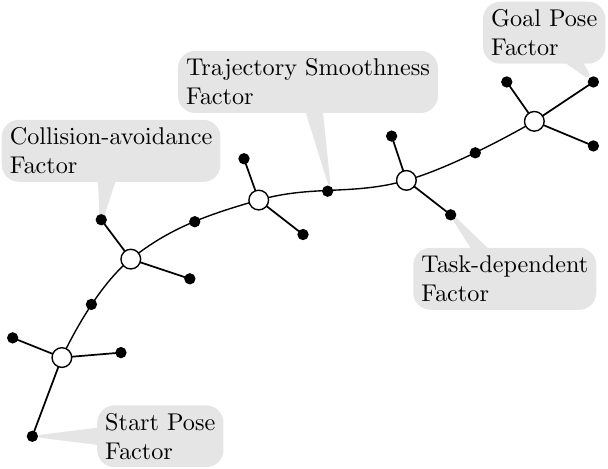}
	\caption{Motion planning factor graph, adapted from~\cite{Mukadam18ijrr_gpmp2}.}
	\label{fig:gpmp2_fg}
\end{figure}

Given the cost functions associated with the constraints mentioned above, the trajectory optimization problem is formulated as follows:
\begin{equation}\label{eq:opt}
\begin{aligned}
x^* &=\argmin_x\{\|h_1(x_s)\|_{\Sigma_1}^2 + \|h_2(x_g)\|_{\Sigma_2}^2 + \\
& \sum_i (\|h_3(x_i)\|_{\Sigma_3}^2 + \|h_4(x_i)\|_{\Sigma_4}^2
+ \|h_5(x_{i-1}, x_i)\|_{\Sigma_5}^2)
\}.
\end{aligned}
\end{equation}
where the $\Sigma$ terms specify the covariance for each cost function in the optimization. A smaller $\Sigma$ value imposes larger penalty for the constraint. Hence, the optimization result will tend to satisfy such a constraint at a higher priority. Eq.~\ref{eq:opt} can be represented as the factor graph shown in Fig.~\ref{fig:gpmp2_fg}, and solved numerically by using iterative approaches such as Gaussian-Newton or Levenberg-Marquardt. In GPMP2, sparsity is exploited through use of the GTSAM solver~\cite{Dellaert12report_gtsam, Dellaert17fnt_fg}.

\subsection{Fast Replanning using the Bayes Tree}
\begin{figure}[!ht]
\centering
\includegraphics[scale=1.1]{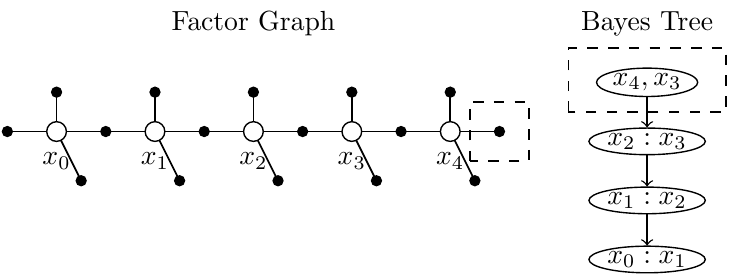}
\caption{An example of motion planning factor graph and the corresponding Bayes tree, adapted from~\cite{Mukadam18ijrr_gpmp2}.
Here we use the goal change of the planning problem as an example to illustrate the idea of incremental replanning with the Bayes tree, where the change of factors in the graph marked with the dashed box only affects part of the Bayes tree as shown in the dashed box.}
\label{fig:bt1}
\end{figure}
Replanning problems are very common in real-world tasks, for instance, when the goal state for the robot has been changed, or obstacles in the environment have been moved around. If the majority of the problem is not changed, then it will not be necessary to solve the entire problem again. 

GPMP2 maintains the solution as a Bayes tree~\cite{Kaess12ijrr}, which is a directed data structure generated from incremental elimination on the underlying factor graph, and incrementally updates this representation in response to changing conditions. Fig.~\ref{fig:bt1} uses a goal change in a planning problem as an example to illustrate the idea of incremental replanning with the Bayes tree, where the change of factors in the graph only affects part of the Bayes tree as shown in the dashed box. 

\section{Extension to Kinodynamic Motion Planning}
When fulfilling both kinematic and dynamic constraints simultaneously, the kinodynamic planning problem becomes more difficult. However, we can formulate the planning problem using factor graphs with the addition of the dynamic constraints, and solve the problem efficiently with numerical optimization tools such as GTSAM, which is able to explore the sparsity in factor graphs, and use heuristics such as minimum-degree MD~\cite{George81book} or nested dissection (ND)~\cite{George73siam} to generate elimination orders which reduce the computation complexity in solving the optimization problem~\cite{Dellaert17fnt_fg}.
\subsection{Manipulator kinodynamic Constraints}
The manipulator kinodynamic constraints include both equality and inequality constraints. For equality constraints, we closely follow~\cite{Lynch17book_robotics, Xie19arxiv_dynamicsFactorGraph}. The following equations of motion express the equality constraints between link $j$ and link $j-1$ imposed by joint $j$, assuming rotational joints.
\newcommand{\V}{\mathcal{V}}
\newcommand{\Vdot}{\mathcal{\dot{V}}}
\newcommand{\G}{\mathcal{G}}
\newcommand{\F}{\mathcal{F}}
\newcommand{\qdot}{\dot{q}}
\newcommand{\qddot}{\ddot{q}}
\newcommand{\Axis}{\mathcal{A}}
\begin{align} 
\label{eq:twist}
&\V_j = [Ad_{T_{j,j-1}(q_j)}]\V_{j-1} - \Axis_j\qdot_{j-1} \\
\label{eq:accel}
&\Vdot_{j} = [Ad_{T_{j,j-1}(q_j)}]\Vdot_{j-1} - \Axis_j\qddot_{j-1} - [ad_{\V_j}]\Axis_j\qdot_{j-1}  \\
\label{eq:wrench}
&Ad^T_{T_{j+1,j}(q_{j+1})}\F_{j+1} = \F_j + \G_j\dot{\V}_j - [ad_{\V_j}]^T\G_j\V_j \\
\label{eq:torque}
&\F_j^T\Axis_j = \tau_j
\end{align}
where $\Axis_j$ (expressed in link $j$ coordinate frame) is the screw axis for joint $j$, and $Ad_{T_{j,j-1}(q_j)}$ is the adjoint map of the transform $T_{j,j-1}$ between the two adjacent links. The relationship between twist $\V_j$ and twist $\V_{j-1}$ is described in Equation \eqref{eq:twist} where $\qdot_{j-1}$ represents the angular velocity of joint $j-1$. 
Equation \eqref{eq:accel} shows the constraint between acceleration $\Vdot_{j-1}$ of link $j-1$ and acceleration $\Vdot_{j}$ of link $j$.
Equation \eqref{eq:wrench} describes the balance between the wrench $\F_j$  and the wrench $\F_{j+1}$ applied to link $j$.
The torque applied at joint {j} is expressed as the projection of wrench $\F_j$ on the screw axis $\Axis_j$ corresponding to joint $j$.

By representing all the variables in the equations of motion at time $t_i$ with state vector $x_i$, we write the cost function corresponding to the above equality constraints in a compact form as the following:  
\begin{align} 
&h_6(x_i) = LHS(EoM(x_i)) - RHS(EoM(x_i))
\end{align}
where $EoM()$ is used to represent the equations of motion from \eqref{eq:twist} to \eqref{eq:torque}, and LHS \textit{and} RHS stand for "Left Hand Side" and "Right Hand Side" of the equations respectively.

Inequality constraints include joint limits such as limits on joint angles, angular velocities, angular accelerations, and torques applied at each joint. We use a hinge loss function to represent the i-th inequality constraint $f(x_i) \leq e_i$, and the cost function $h_7(x_i)$ under the current state can be written as
\begin{equation}\label{eq:inequality}
h_7(x_i) = c_{limit}(f(x_i))
\end{equation}
where $c_{limit}()$ is defined as the hinge loss:
\begin{equation}\label{eq:hinge}
c_{limit}(z)=\left\{
\begin{array}{ll}
a(z_l - z + \varepsilon) &{if \ z - z_l \leq \varepsilon}\\
a(z - z_u + \varepsilon) &{if \ z_u - z \leq \varepsilon}\\
0\ &{o/w}
\end{array}\right.
\end{equation}
where $z_l$ is the lower limit, $z_u$ is the upper limit, $\varepsilon$ is a constant threshold to prevent exceeding the limit, and $a$ is a constant ratio which determines how fast the error grows as the value approaches the limit. If the value is not within the threshold, then there will be no cost. In this way, the limit violations are prevented during the optimization. 


The DFG of an RR manipulator shown in Fig.~\ref{fig:dfg} is used to illustrate the dynamic constraints in kinodynamic motion planning.
This DFG associated with the motion planning considers the robot dynamics, and ensures that all dynamic constraints are satisfied when solving for a motion plan.

\subsection{Task-dependent Dynamic Constraints}
In addition, the planner sometimes will have to optimize a user-specified cost function such as a minimum torque constraint, for which the cost function can be expressed as
\begin{equation}\label{eq:task}
h_8(x_i) = \sum_j^N\tau_j(x_i)
\end{equation}
where $\tau_j(x_i)$ is the torque at joint j expressed as a function of state vector $x_i$, which can be calculated through inverse dynamics. We can directly retrieve all the torque values from the DFG.


\subsection{Trajectory Smoothness}
Here we use a continuous-time configuration space trajectory with a constant acceleration instead of a constant velocity, where the state transition matrix $\Phi(t_i,t_{i-1})$ in equation \eqref{eq:smoothness} is replaced with
\begin{equation}\label{eq:phi_accel}
\Phi(t_i,t_{i-1}) = \begin{bmatrix} 1 & \Delta t_i & \frac{1}{2}(\Delta t_i)^2 \\ 0 & 1 & \Delta t_i \\ 0 & 0 & 1 \end{bmatrix}
\end{equation}
and the covariance matrix is 
\begin{equation}\label{eq:sigma_accel}
\Sigma_5 = \begin{bmatrix} \frac{1}{2} \Delta t_i^5 \mathbf{Q}_C &
\frac{1}{8} \Delta t_i^4 \mathbf{Q}_C & \frac{1}{6} \Delta t_i^3 \mathbf{Q}_C\\ \frac{1}{8} \Delta t_i^4 \mathbf{Q}_C & 
\frac{1}{3} \Delta t_i^3 \mathbf{Q}_C & \frac{1}{2} \Delta t_i^2 \mathbf{Q}_C \\ \frac{1}{6} \Delta t_i^3 \mathbf{Q}_C & \frac{1}{2} \Delta t_i^2 \mathbf{Q}_C &
\Delta t_i \mathbf{Q}_C \end{bmatrix}
\end{equation}

\subsection{Kinodynamic Motion Planning Factor Graph}
A factor graph is used to solve the kinodynamic motion planning problem, where constraints including start and goal pose constraints, manipulator dynamic constraints, task-dependent kinematic and dynamic constraints, and constraints that ensure a smooth and collision-free trajectory are represented as factors in the graph, as shown in Fig.~\ref{fig:dfgp}. 

\section{Fast Replanning with Incremental Technique}
\begin{figure*}[!htb]
	\centering
	\begin{subfigure}[b]{0.25\textwidth}
		\centering
		\includegraphics[scale=0.25]{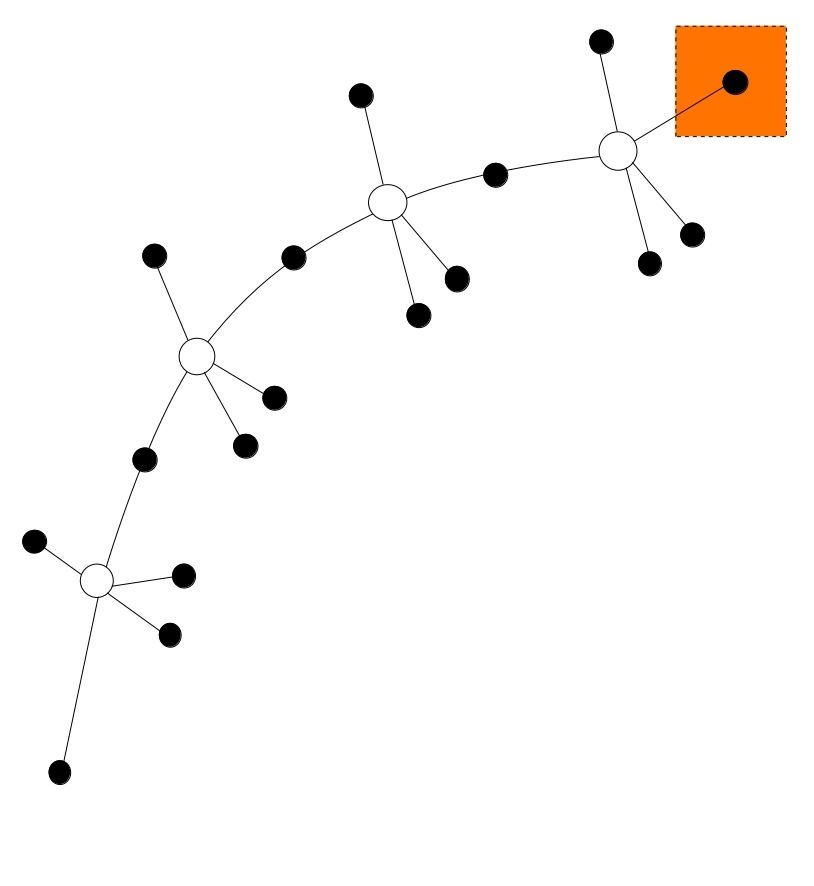}
		\caption{}
		\label{fig:fg}
	\end{subfigure}
	\begin{subfigure}[b]{0.65\textwidth}
		\centering
		\includegraphics[scale=0.25]{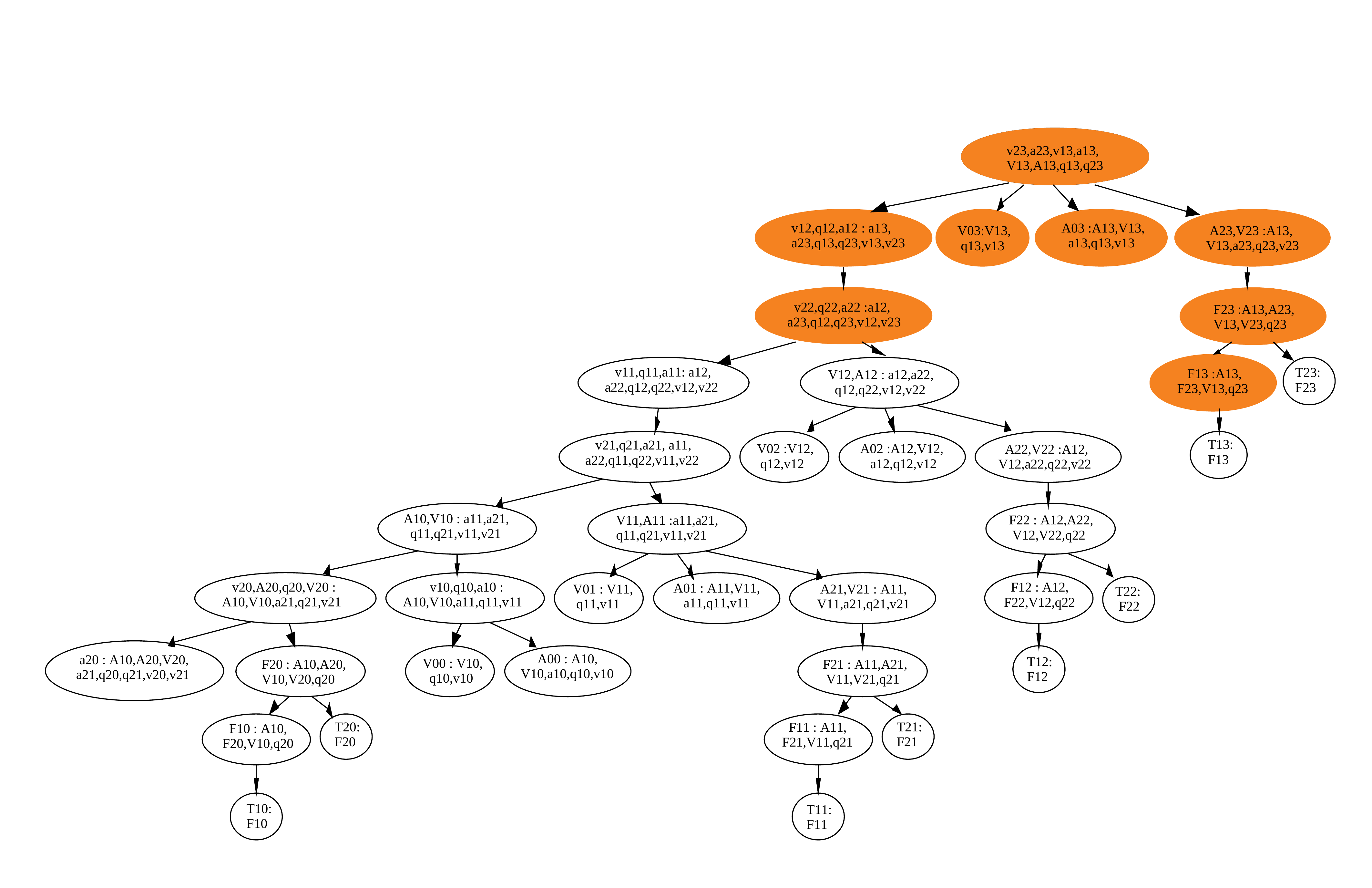}
		\caption{}
		\label{fig:bt}
	\end{subfigure}
	\caption{(a) An example factor graph for a kinodynamic motion replanning problem of a robot with two links and two revolute joints, where the orange shaded regions show the factor changed due to a change of the goal configuration; (b) The corresponding Bayes tree with the orange shaded regions showing the parts affected by the change of factors in the graph. In this Bayes tree, the characters q, v, a, V, A, F and T represent joint angle, joint velocity, joint acceleration, link twist, link acceleration, wrench, and torque respectively. The first number ranging from 1 to 2 represents the link or joint index, and the second number ranging from 0 to 3 represents the time index.}
	\label{fig:btfg}
\end{figure*}

We are able to solve kinodynamic motion replanning problems efficiently by taking advantage of the incremental technique with iSAM2, which uses the Bayes tree to perform incremental optimization on the factor graph. In the following kinodynamic motion replanning problem, a RR manipulator needs to find a minimum torque trajectory to move the end-effector from the given start pose to the desired goal pose while respecting both kinematic and dynamic constraints. A Bayes tree is generated by eliminating the variables in the kinodynamic motion planning factor graph from the first state to the last state. With a change of the goal configuration, only the factor in the orange shaded portion of the graph shown in Fig.~\ref{fig:btfg} will be changed, and the new planning problem can be solved with the corresponding Bayes tree, where the orange shaded regions show the affected parts compared to the original Bayes tree. The optimization result obtained from the Bayes tree will only be partially changed since the majority of the Bayes tree is left unchanged.

The proposed kinodynamic motion replanning algorithm is implemented with iSAM2 incremental solver~\cite{Kaess12ijrr}. First, we check for the information which has been updated; next, we change the factors in the graph according to the updated information; finally, we run the incremental algorithm to update the Bayes tree implemented in iSAM2 to get the new solution.


\section{Evaluation}
\subsection{Experiment Setup}
We implemented both a batch version (DFGP) and an incremental version (iDFGP) of the method above using the GTSAM C++ library~\cite{Dellaert12report_gtsam,Dellaert17fnt_fg}, and ran simulations in V-REP for visualization. We used the Levenberg-Marquardt algorithm to solve the nonlinear least squares optimization problem for KMDP, with an initial value for $\lambda = 0.01$, and terminating the optimization process if either of the following conditions is satisfied: 1) it reaches a maximum of 200 iterations, or 2) the relative decrease in error is smaller than $10^{-5}$. For the incremental version, iDFGP uses the iSAM2~\cite{Kaess12ijrr} optimizer with default settings.

\subsection{Kinodynamic Motion Planning for the Acrobot}
\begin{figure}[!ht]
	\centering
	\includegraphics[scale=0.5]{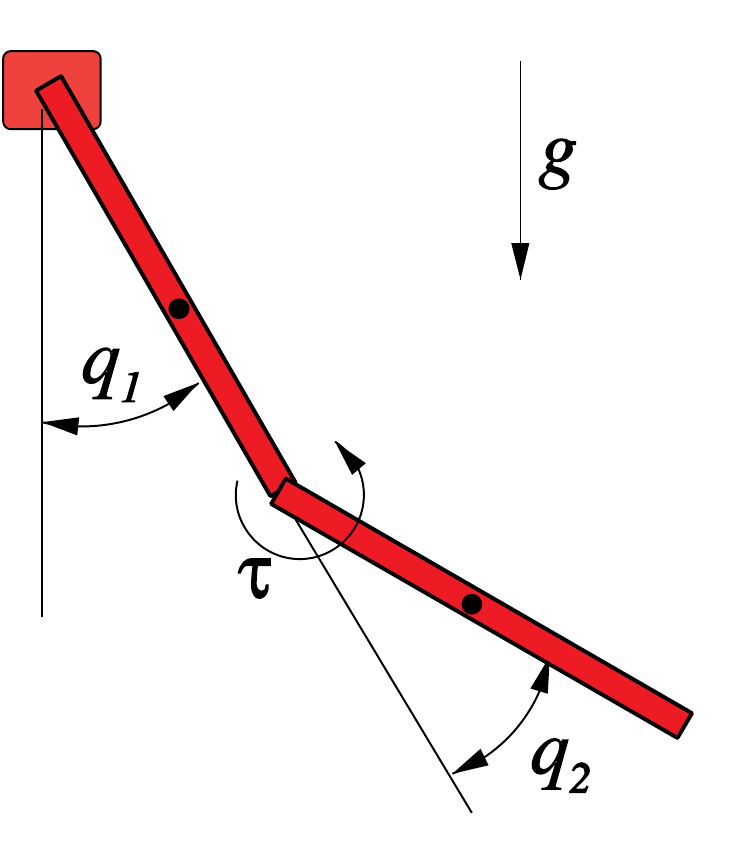}
	\caption{Acrobot}
	\label{fig:Acrobot}
\end{figure}

\begin{table*}[!htb]
	\caption{Results for Kinodynamic Motion Planning.}
	\label{table:results_DFGP}
	\begin{center}
		\begin{tabular}{| c | l | c | c | c | c |}
			\hline 
			\multicolumn{2}{|c|}{Tests} & DFGP & SST & SST* & AO-RRT \\
			\hline
			\multirow{3}*{Acrobot}     
			& Success Rate (\%) & 100 & 98 & 100 & 90 \\
			& Average Time (s) & 1.07 & 180.55 & 229.67 & 1058.48 \\
			& Maximum Time (s) & 1.07 & 1054.89 & 2152.95 & 5868.87\\
			\hline
			\multirow{3}*{Kuka Task1}     
			& Success Rate (\%) & 86 & -- & -- & -- \\
			& Average Time (s) & 2.15 & -- & -- & -- \\
			& Maximum Time (s) & 3.39 & -- & -- & --\\
			\hline 
			\multirow{3}*{Kuka Task2}     
			& Success Rate (\%) & 90 & -- & -- & -- \\
			& Average Time (s) & 1.75 & -- & -- & --\\
			& Maximum Time (s)   & 2.92 & -- & -- & --\\
			\hline 
		\end{tabular}
	\end{center}
\end{table*}

\begin{figure*}[!htb]
	\centering
	\begin{subfigure}[b]{0.09\textwidth}
		\centering
		\includegraphics[scale=0.12]{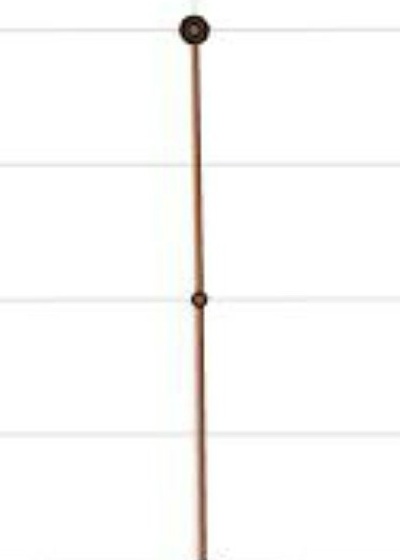}
		\caption{}
		\label{fig:acrobot6fg1}
	\end{subfigure}
	\begin{subfigure}[b]{0.09\textwidth}
		\centering
		\includegraphics[scale=0.12]{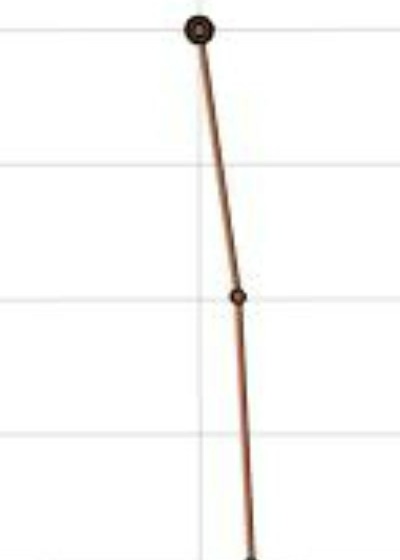}
		\caption{}
		\label{fig:acrobot6fg2}
	\end{subfigure}
	\begin{subfigure}[b]{0.09\textwidth}
		\centering
		\includegraphics[scale=0.12]{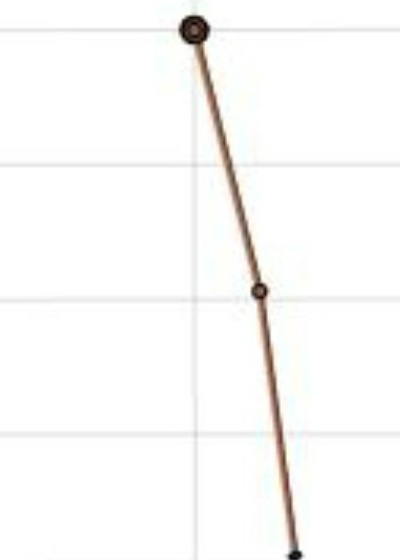}
		\caption{}
		\label{fig:acrobot6fg3}
	\end{subfigure}
	\begin{subfigure}[b]{0.09\textwidth}
		\centering
		\includegraphics[scale=0.12]{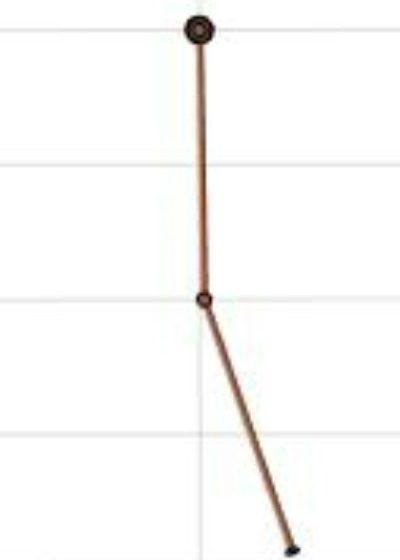}
		\caption{}
		\label{fig:acrobot6fg4}
	\end{subfigure}
	\begin{subfigure}[b]{0.09\textwidth}
		\centering
		\includegraphics[scale=0.12]{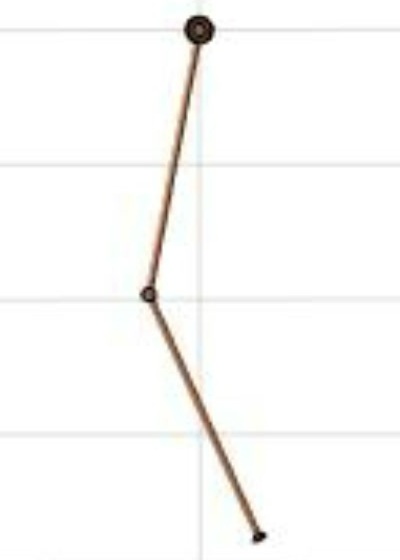}
		\caption{}
		\label{fig:acrobot6fg5}
	\end{subfigure}
	\begin{subfigure}[b]{0.09\textwidth}
		\centering
		\includegraphics[scale=0.12]{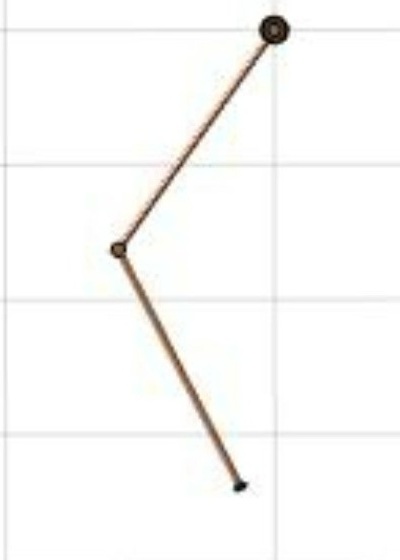}
		\caption{}
		\label{fig:acrobot6fg6}
	\end{subfigure}
	\begin{subfigure}[b]{0.09\textwidth}
		\centering
		\includegraphics[scale=0.12]{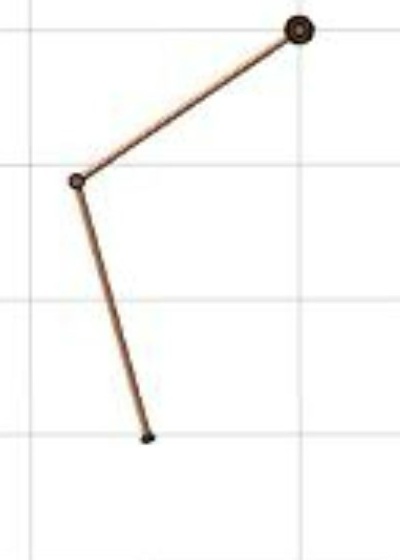}
		\caption{}
		\label{fig:acrobot6fg7}
	\end{subfigure}
	\begin{subfigure}[b]{0.09\textwidth}
		\centering
		\includegraphics[scale=0.12]{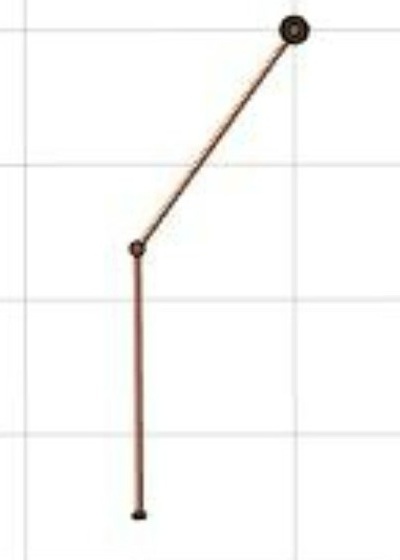}
		\caption{}
		\label{fig:acrobot6fg8}
	\end{subfigure}
	\begin{subfigure}[b]{0.09\textwidth}
		\centering
		\includegraphics[scale=0.12]{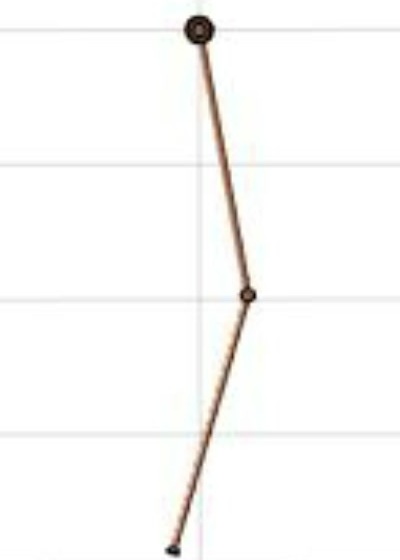}
		\caption{}
		\label{fig:acrobot6fg9}
	\end{subfigure}
	\begin{subfigure}[b]{0.09\textwidth}
		\centering
		\includegraphics[scale=0.12]{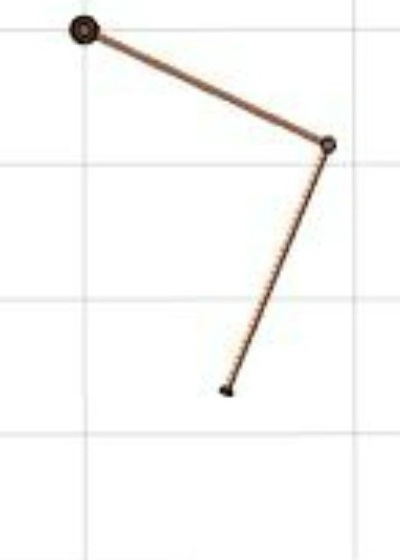}
		\caption{}
		\label{fig:acrobot6fg10}
	\end{subfigure}
	\begin{subfigure}[b]{0.09\textwidth}
		\centering
		\includegraphics[scale=0.12]{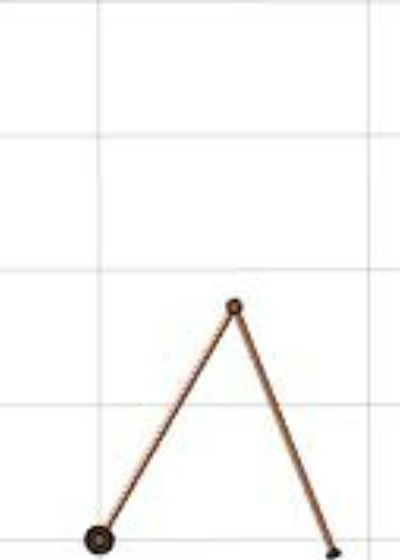}
		\caption{}
		\label{fig:acrobot6fg11}
	\end{subfigure}
	\begin{subfigure}[b]{0.09\textwidth}
		\centering
		\includegraphics[scale=0.12]{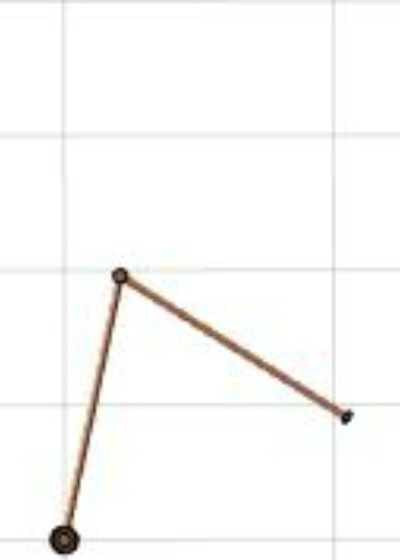}
		\caption{}
		\label{fig:acrobot6fg12}
	\end{subfigure}
	\begin{subfigure}[b]{0.09\textwidth}
		\centering
		\includegraphics[scale=0.12]{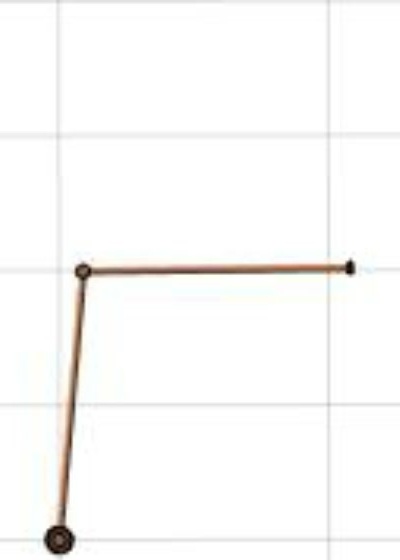}
		\caption{}
		\label{fig:acrobot6fg13}
	\end{subfigure}
	\begin{subfigure}[b]{0.09\textwidth}
		\centering
		\includegraphics[scale=0.12]{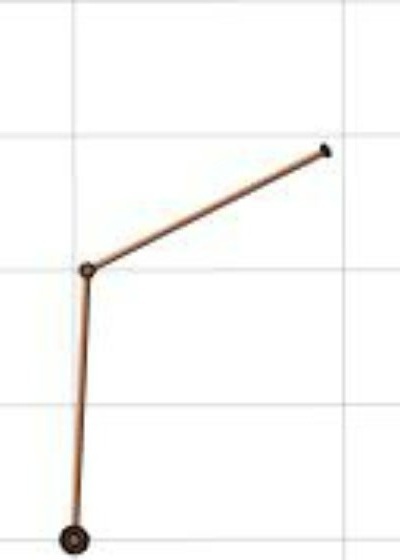}
		\caption{}
		\label{fig:acrobot6fg14}
	\end{subfigure}
	\begin{subfigure}[b]{0.09\textwidth}
		\centering
		\includegraphics[scale=0.12]{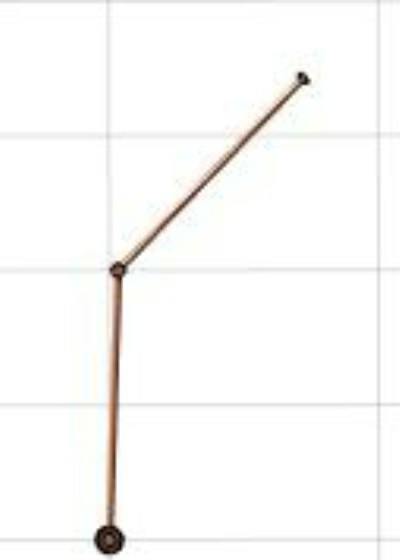}
		\caption{}
		\label{fig:acrobot6fg15}
	\end{subfigure}
	\begin{subfigure}[b]{0.09\textwidth}
		\centering
		\includegraphics[scale=0.12]{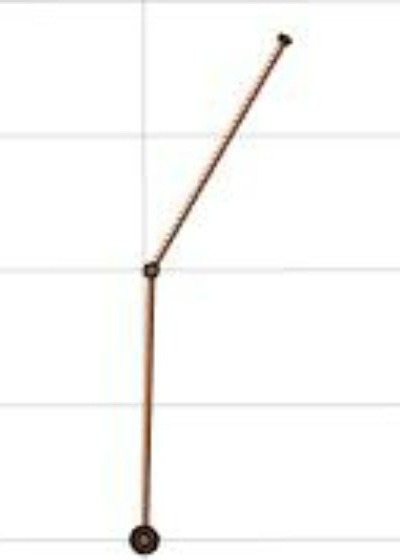}
		\caption{}
		\label{fig:acrobot6fg16}
	\end{subfigure}
	\begin{subfigure}[b]{0.09\textwidth}
		\centering
		\includegraphics[scale=0.12]{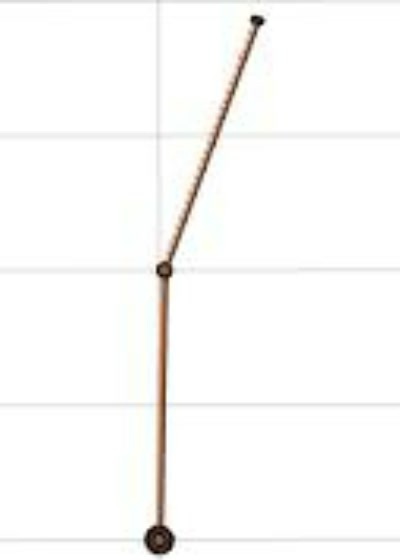}
		\caption{}
		\label{fig:acrobot6fg17}
	\end{subfigure}
	\begin{subfigure}[b]{0.09\textwidth}
		\centering
		\includegraphics[scale=0.12]{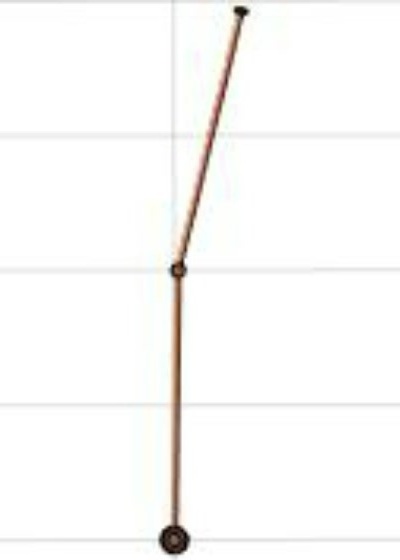}
		\caption{}
		\label{fig:acrobot6fg18}
	\end{subfigure}
	\begin{subfigure}[b]{0.09\textwidth}
		\centering
		\includegraphics[scale=0.12]{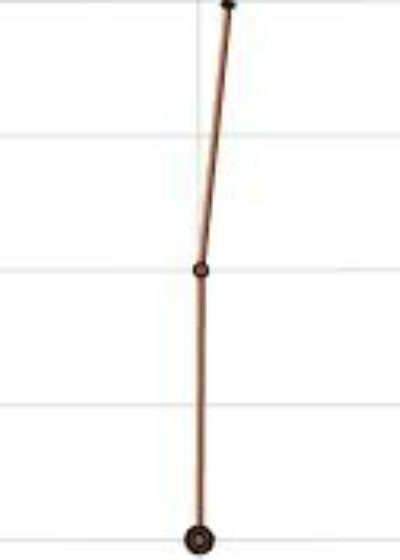}
		\caption{}
		\label{fig:acrobot6fg19}
	\end{subfigure}
	\begin{subfigure}[b]{0.09\textwidth}
		\centering
		\includegraphics[scale=0.12]{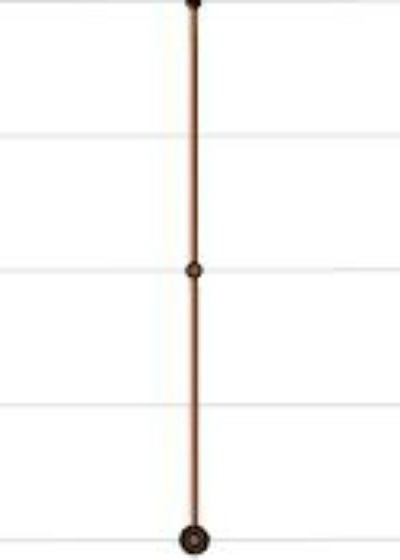}
		\caption{}
		\label{fig:acrobot6fg20}
	\end{subfigure}
	\caption{A solution for the kinodynamic motion planning problem where the task is to bring the Acrobot from the rest configuration as in (a) to the goal configuration as in (t). Images (b)-(s) show the intermediate configurations along the planned trajectory.}
	\label{fig:acrobot6fg}
\end{figure*}

The Acrobot~\cite{Tedrake09book_underactuatedRoboitc} is a planar RR manipulator with an actuator at the elbow but no actuator at the shoulder, as shown in Fig.~\ref{fig:Acrobot}. The task is to bring the Acrobot from its initial rest configuration to the upright pose where there is limited torque applied at the elbow joint. The equations of motion of the Acrobot are described in~\cite{Tedrake09book_underactuatedRoboitc}. 

We perform kinodynamic motion planning for the Acrobot with our DFGP algorithm. Due to the lack of source code for any other optimization-based kinodynamic motion planning algorithms, we only benchmark it against some of the state-of-art sampling-based kinodynamic motion planners, such as Stable-Sparse-RRT (SST), Stable-Sparse-RRT* (SST*) and Asymptotically-Optimal-RRT (AO-RRT). All benchmarks were run on a single thread of a 4.0GHz Intel Core i7 CPU.

We used the implementations of the RRT-style methods from an open source repository~\cite{rrt} and implemented a test case for the Acrobot using the framework of the example problem in the repository. The parameters used for SST and SST* are suggested by the author of the original paper~\cite{Li16ijrr_kinoSample}, and default settings are used for AO-RRT. We ran each of the RRT algorithms 50 times, and each time the test was terminated if any of the following events occurred: 1) found a solution with trajectory time length less than 10 seconds, or 2) reached a maximum number of 300k iterations. We counted a trial as a success if a trajectory with a time less or equal to 10 seconds was returned, and calculated the success rate and the average computation time based on the successful trials.

For the proposed DFGP algorithm, we set the trajectory length as 10 seconds and initialized the joint trajectory with an acceleration-smooth straight-line, similar to initialization methods used in~\cite{Mukadam16icra_GPMP,Mukadam18ijrr_gpmp2}. We ran DFGP 50 times and calculated the average computation time.

The benchmark results for the Acrobot are presented in Table \ref{table:results_DFGP}. It should not be surprising to find that RRT-style methods are mush slower than the proposed DFGP algorithm. From the results of RRT-style methods, we can see the difficulty of solving the kinodynamic motion planning problem for a under-actuated robotic system. Without taking full dynamics of the Acrobot into consideration, it is almost impossible for a motion planner to find a solution to bring the Acrobot from its initial rest configuration to the upright pose. Fig.~\ref{fig:acrobot6fg} shows a solution of the proposed DFGP algorithm, in which we observe that the Acrobot swings back and forth to gain momentum while bending the first joint to reduce the moment arm so that it can move the end effector to the goal pose.

From basic system dynamics, we know there are only two equilibrium points for the Acrobot; one is at the rest configuration, and the other one is at the upright configuration~\cite{Spong95csm_swing}. Therefore, we did not attempt any replanning experiments in the case of the Acrobot.   

\subsection{Kinodynamic Motion Planning for a KUKA Arm}
\begin{figure*}[!htb]
	\centering
	\begin{subfigure}[b]{0.195\textwidth}
		\centering
		\includegraphics[scale=0.092]{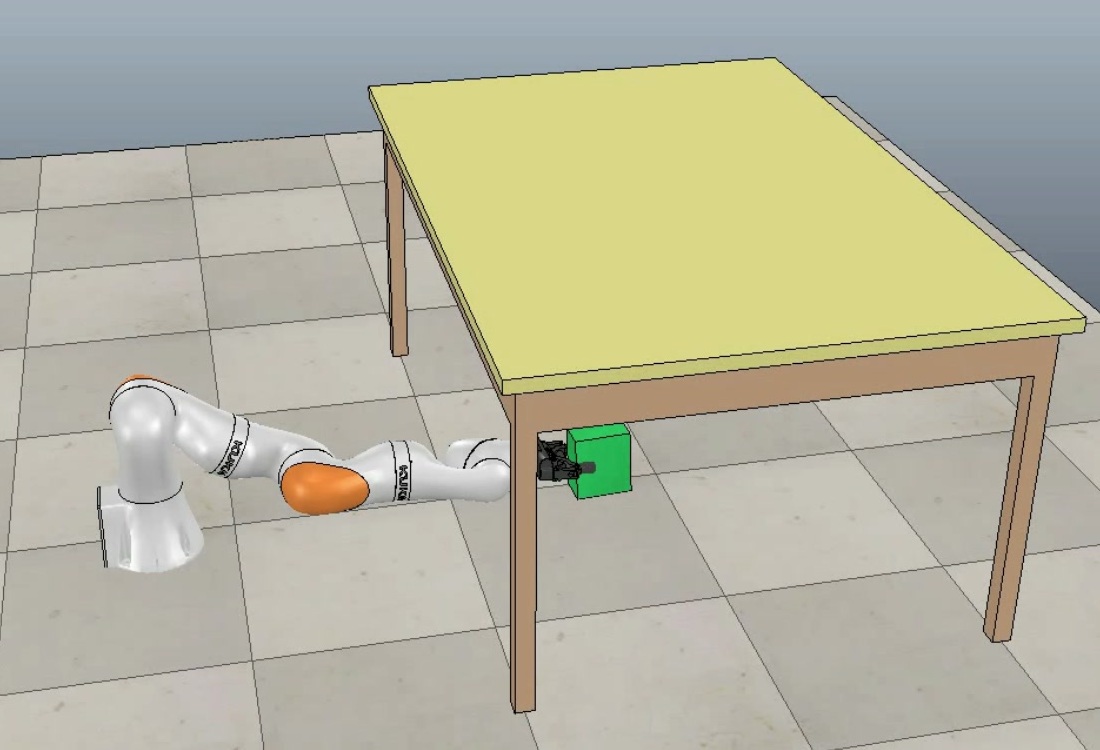}
		\caption{}
		\label{fig:kukaObs0}
	\end{subfigure}
	\begin{subfigure}[b]{0.195\textwidth}
		\centering
		\includegraphics[scale=0.092]{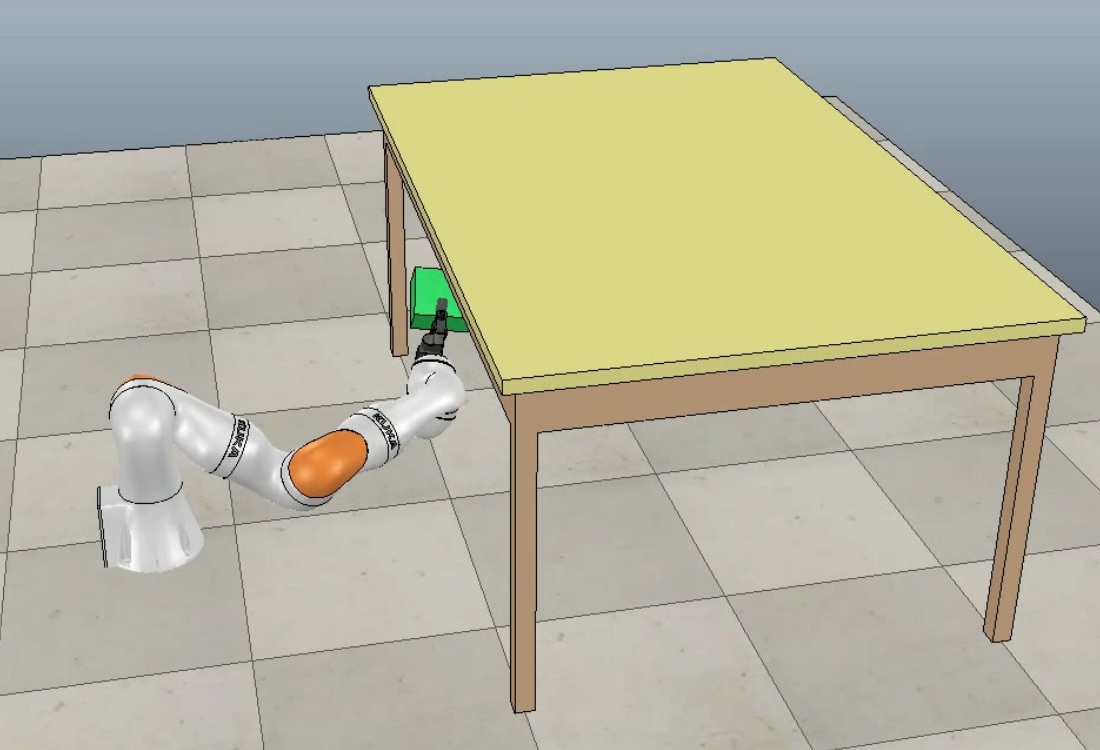}
		\caption{}
		\label{fig:kukaObs1}
	\end{subfigure}
	\begin{subfigure}[b]{0.195\textwidth}
		\centering
		\includegraphics[scale=0.092]{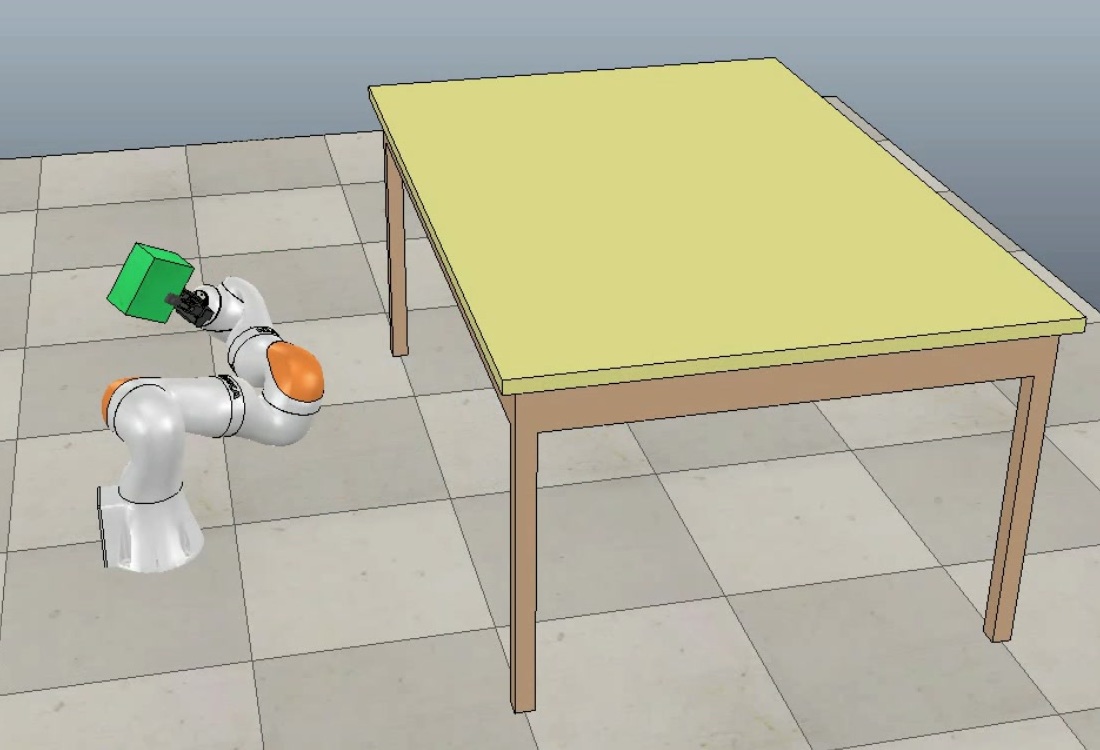}
		\caption{}
		\label{fig:kukaObs2}
	\end{subfigure}
	\begin{subfigure}[b]{0.195\textwidth}
		\centering
		\includegraphics[scale=0.092]{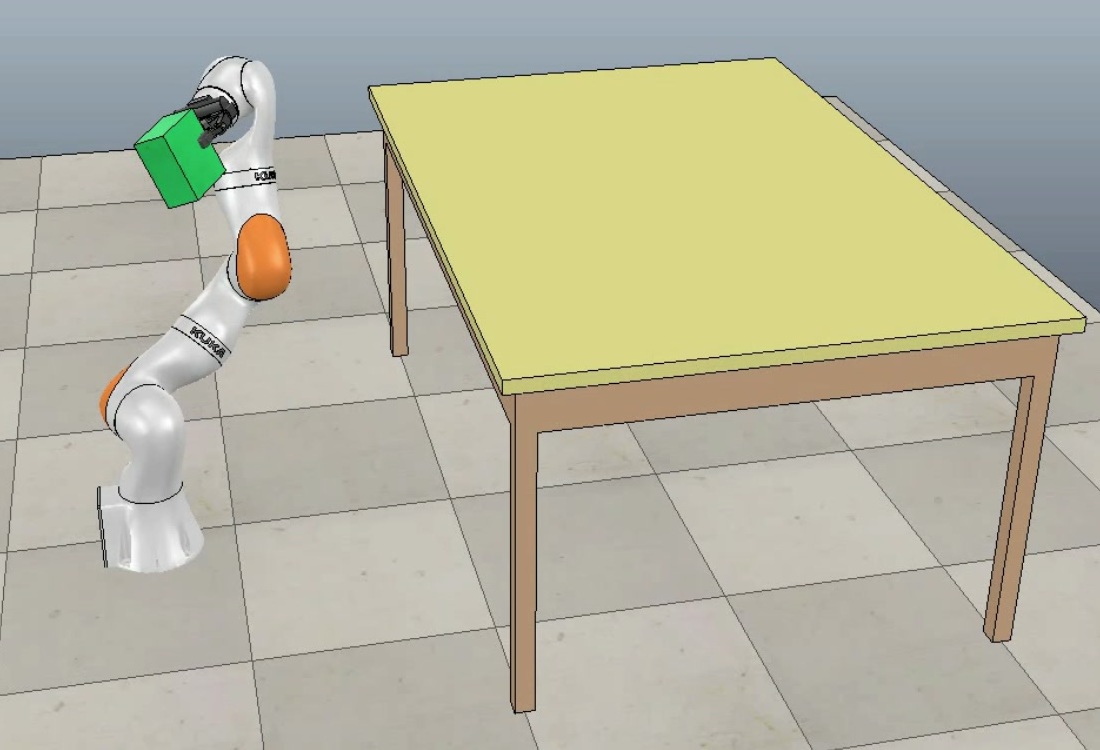}
		\caption{}
		\label{fig:kukaObs3}
	\end{subfigure}
	\begin{subfigure}[b]{0.195\textwidth}
		\centering
		\includegraphics[scale=0.092]{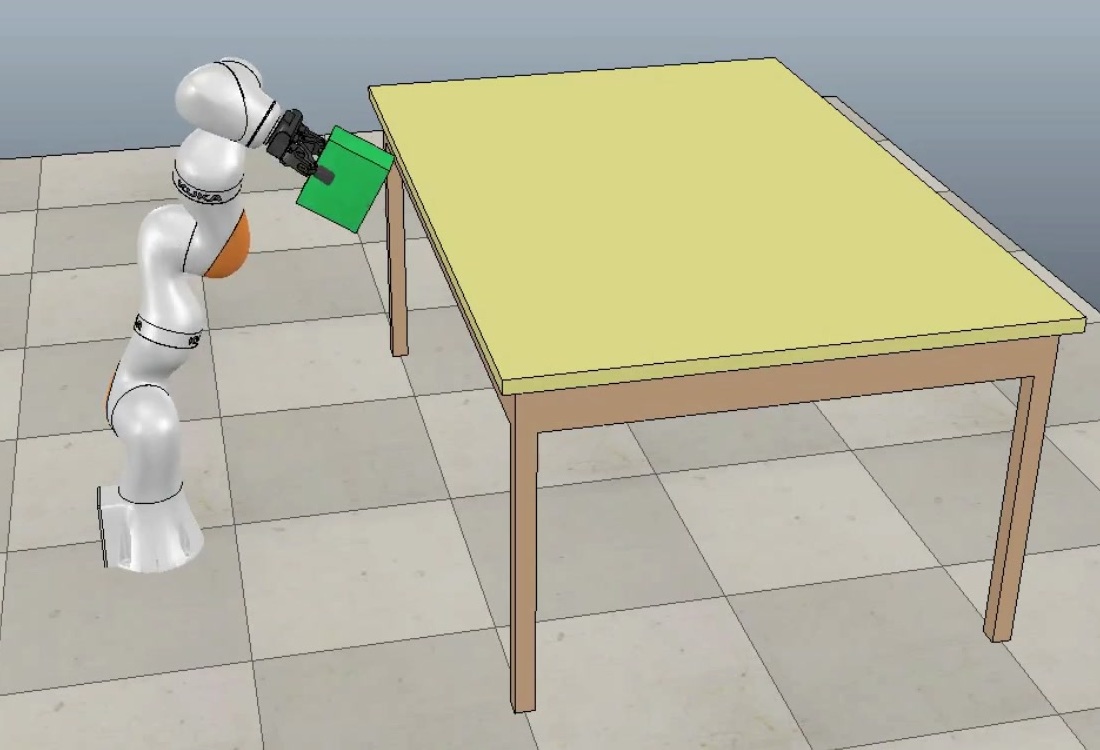}
		\caption{}
		\label{fig:kukaObs4}
	\end{subfigure}
	\begin{subfigure}[b]{0.195\textwidth}
		\centering
		\includegraphics[scale=0.092]{figures/kukaObs0.jpg}
		\caption{}
		\label{fig:rekukaObs0}
	\end{subfigure}
	\begin{subfigure}[b]{0.195\textwidth}
		\centering
		\includegraphics[scale=0.092]{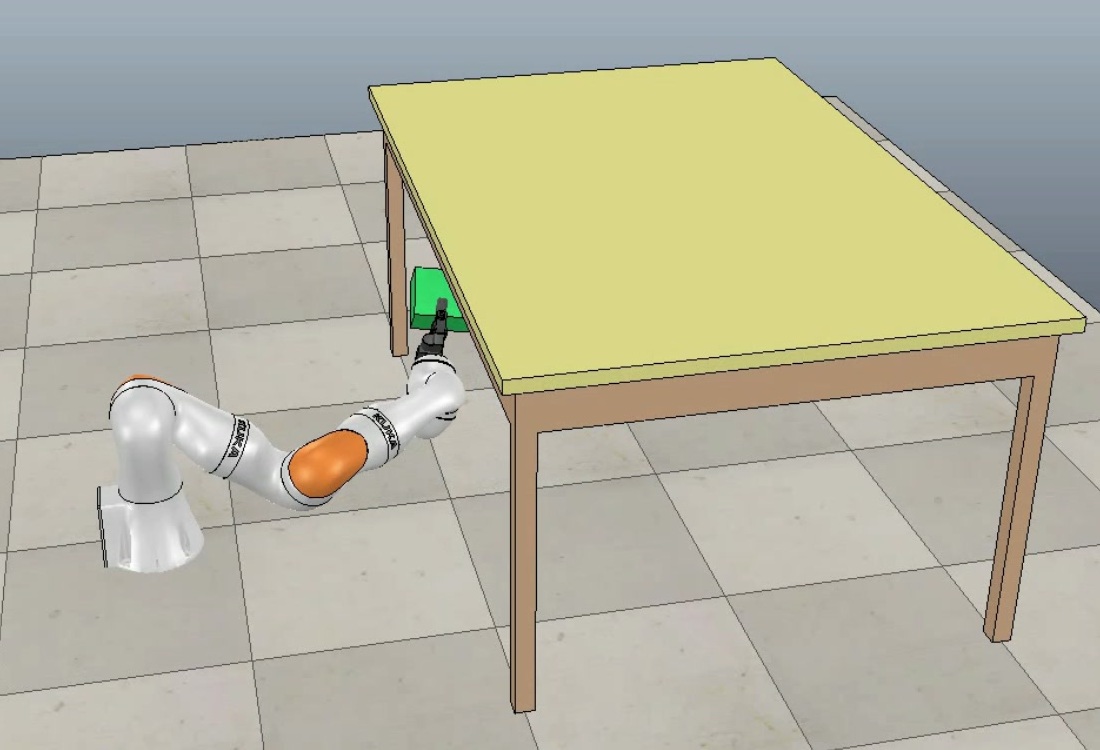}
		\caption{}
		\label{fig:rekukaObs1}
	\end{subfigure}
	\begin{subfigure}[b]{0.195\textwidth}
		\centering
		\includegraphics[scale=0.092]{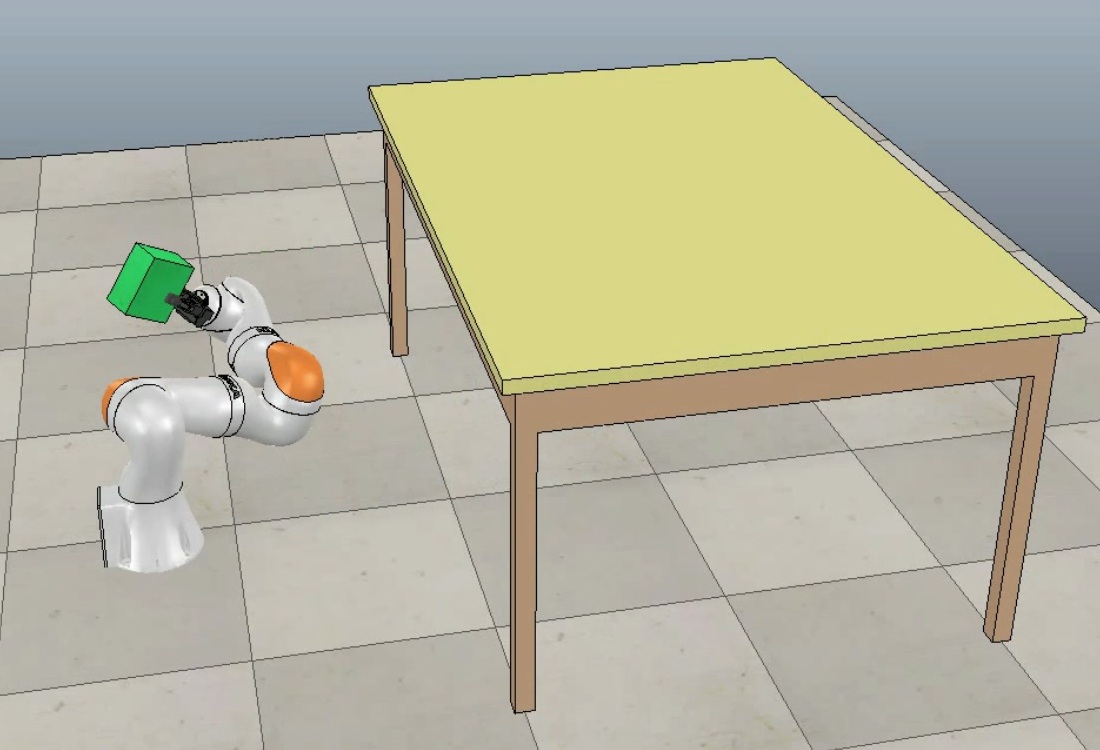}
		\caption{}
		\label{fig:rekukaObs2}
	\end{subfigure}
	\begin{subfigure}[b]{0.195\textwidth}
		\centering
		\includegraphics[scale=0.092]{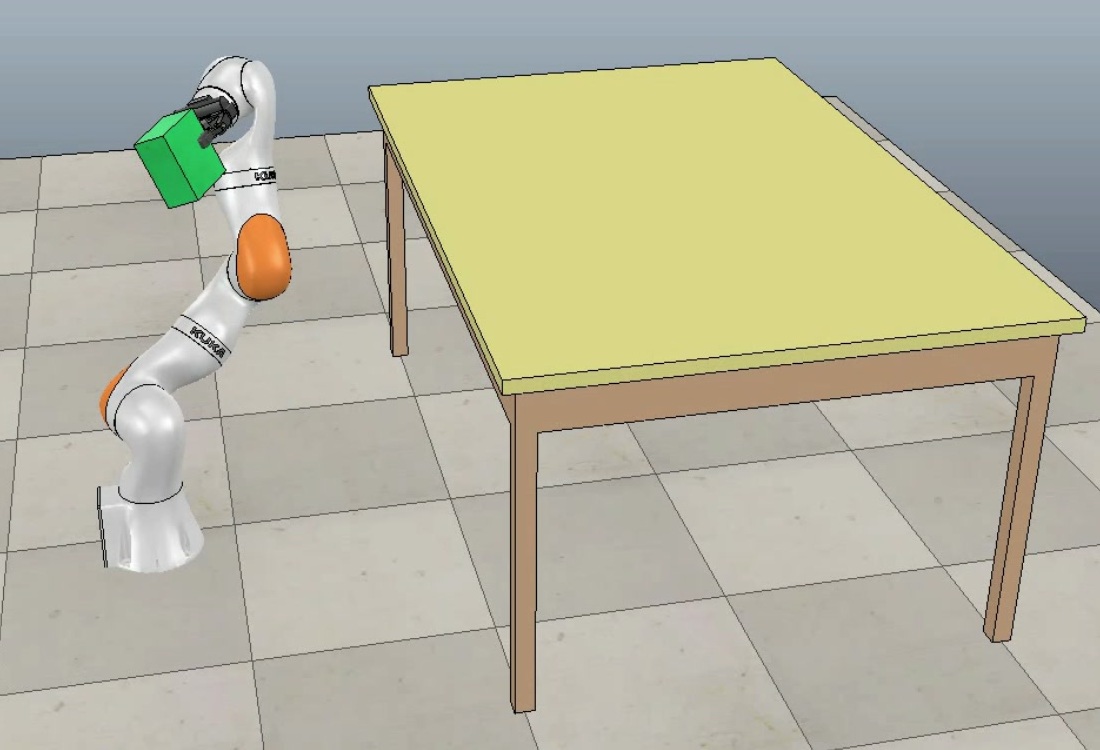}
		\caption{}
		\label{fig:rekukaObs3}
	\end{subfigure}
	\begin{subfigure}[b]{0.195\textwidth}
		\centering
		\includegraphics[scale=0.092]{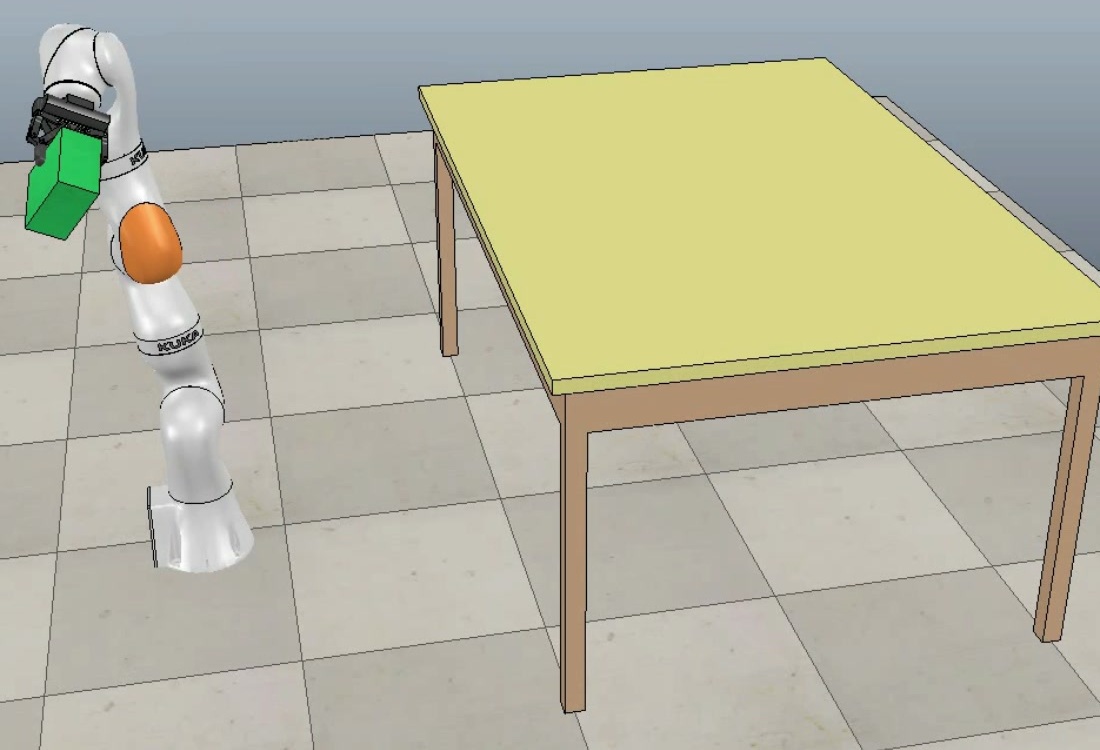}
		\caption{}
		\label{fig:rekukaObs4}
	\end{subfigure}
	\caption{(a)-(e) show a solution obtained by the DFGP algorithm for kinodynamic motion planning task 1, where the objective is to find a trajectory for the KUKA robot to bring a block from the initial location as shown in (a) to the goal location as shown in (e) while avoiding collision. (f)-(j) show a solution obtained by the iDFGP algorithm for the corresponding replanning problem with the new goal as shown in (j).}
	\label{fig:kuka-task1}
\end{figure*}

\begin{figure*}[!htb]
	\centering
	\begin{subfigure}[b]{0.195\textwidth}
		\centering
		\includegraphics[scale=0.092]{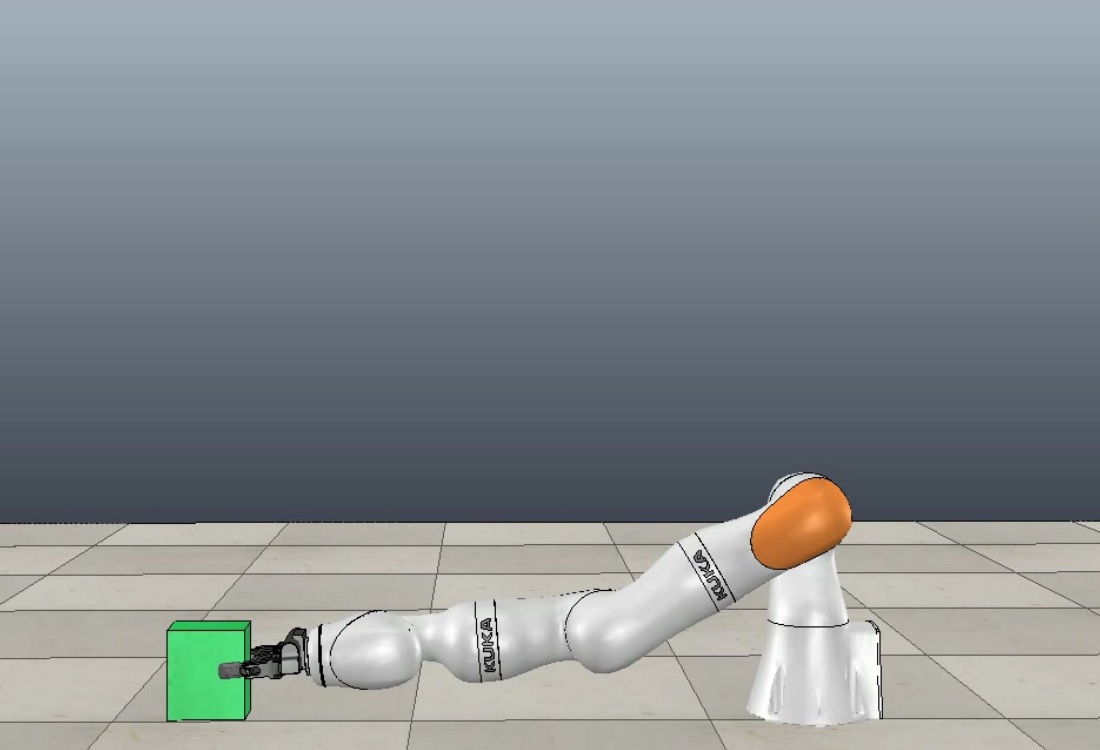}
		\caption{}
		\label{fig:kukaNoMinTorque0}
	\end{subfigure}
	\begin{subfigure}[b]{0.195\textwidth}
		\centering
		\includegraphics[scale=0.092]{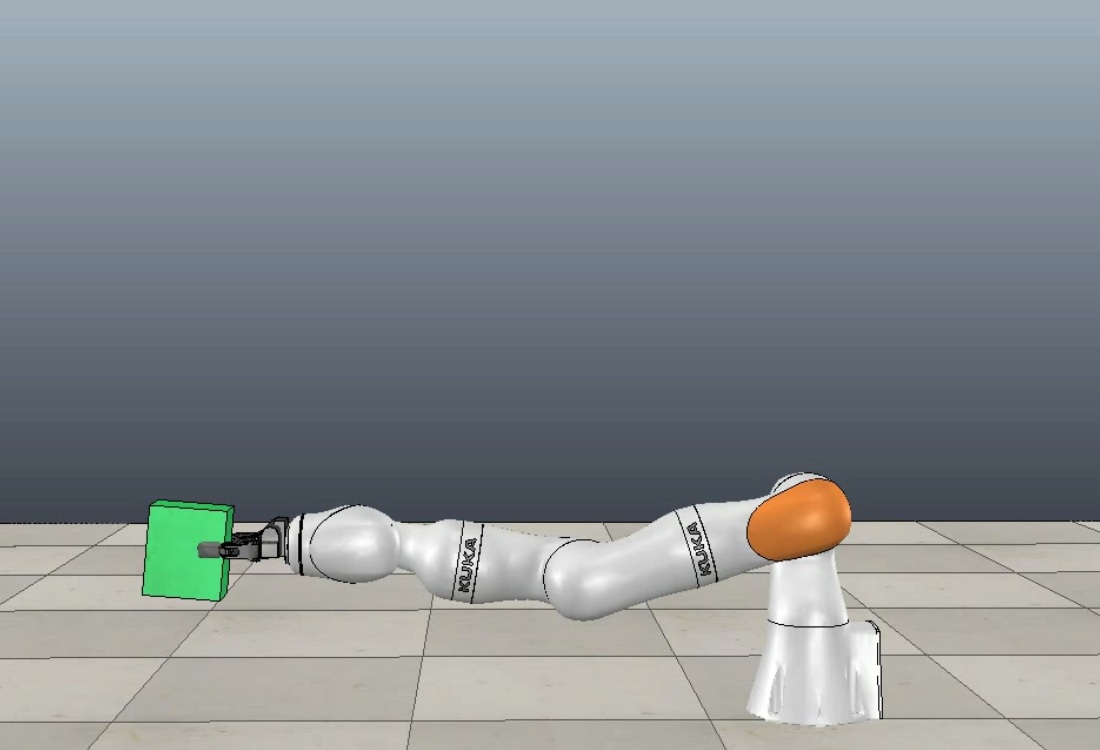}
		\caption{}
		\label{fig:kukaNoMinTorque1}
	\end{subfigure}
	\begin{subfigure}[b]{0.195\textwidth}
		\centering
		\includegraphics[scale=0.092]{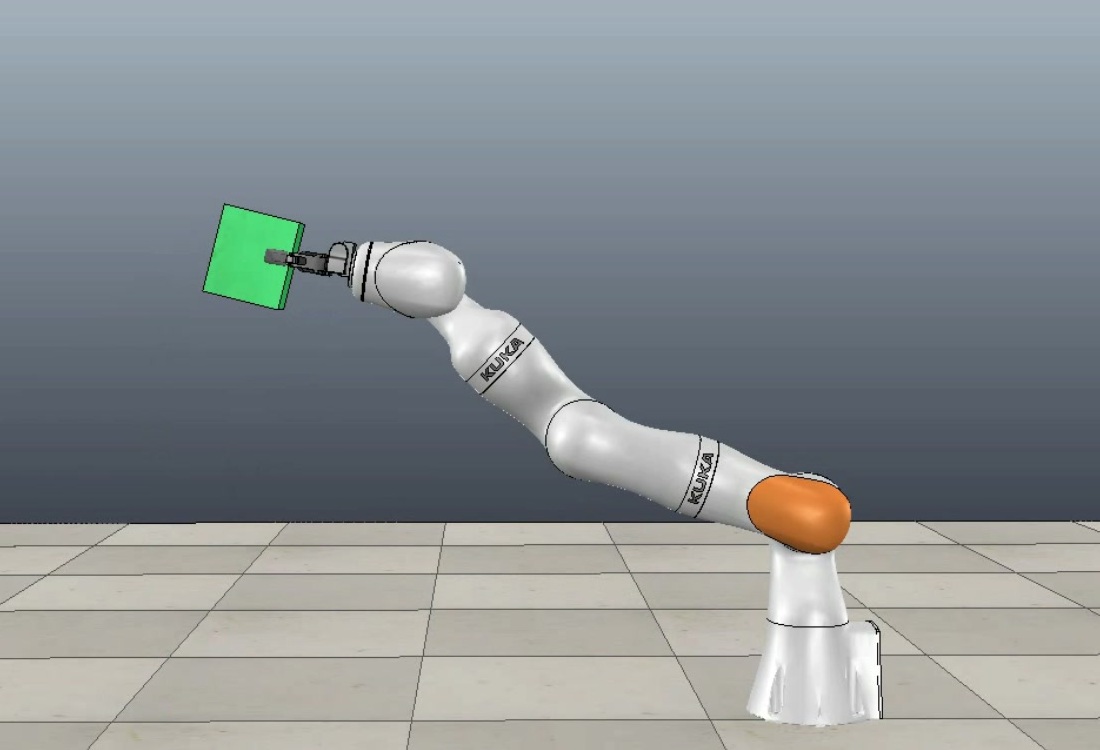}
		\caption{}
		\label{fig:kukaNoMinTorque2}
	\end{subfigure}
	\begin{subfigure}[b]{0.195\textwidth}
		\centering
		\includegraphics[scale=0.092]{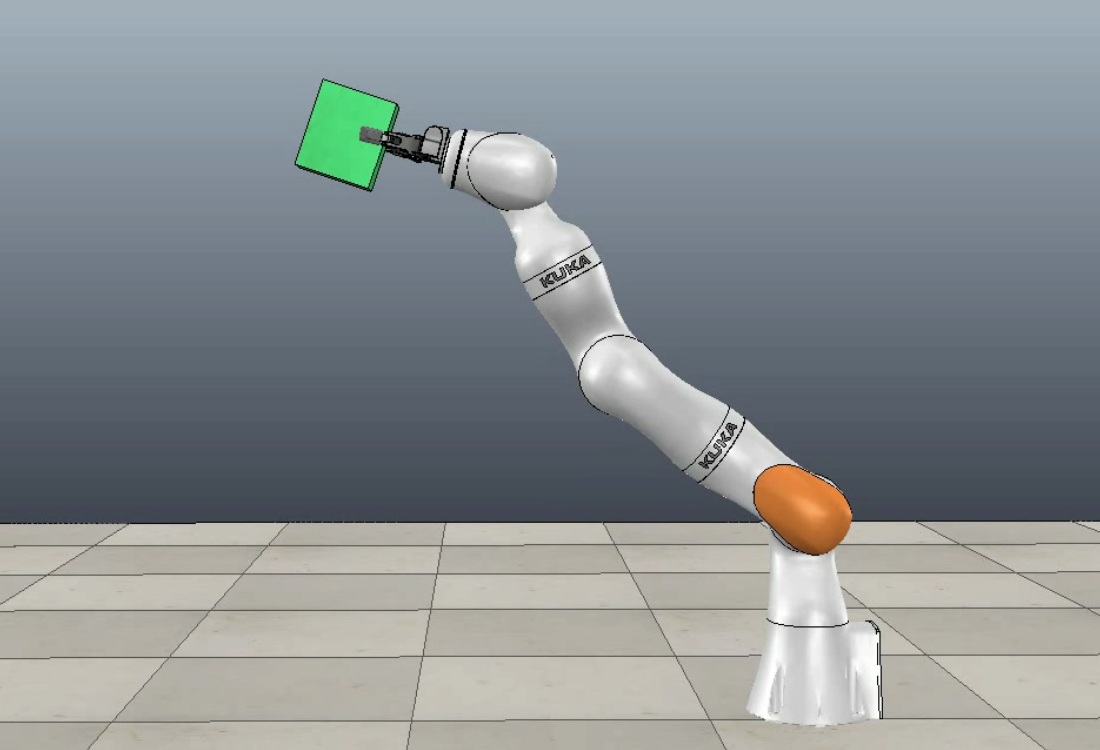}
		\caption{}
		\label{fig:kukaNoMinTorque3}
	\end{subfigure}
	\begin{subfigure}[b]{0.195\textwidth}
		\centering
		\includegraphics[scale=0.092]{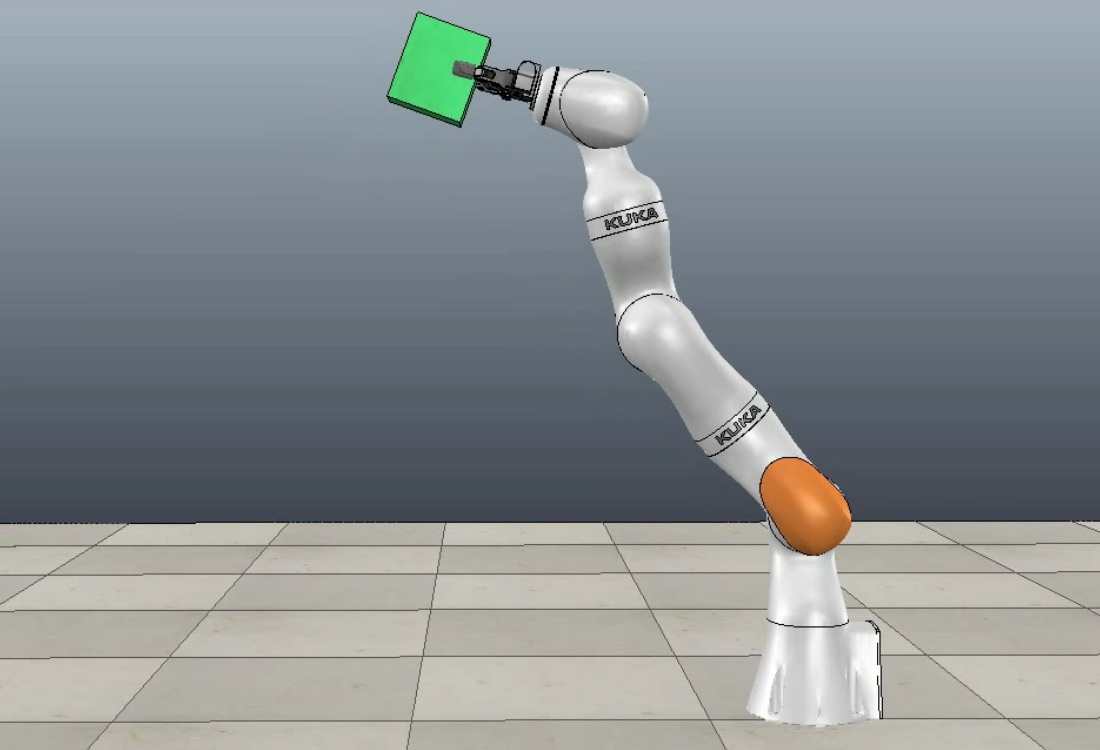}
		\caption{}
		\label{fig:kukaNoMinTorque4}
	\end{subfigure}
	\begin{subfigure}[b]{0.195\textwidth}
		\centering
		\includegraphics[scale=0.092]{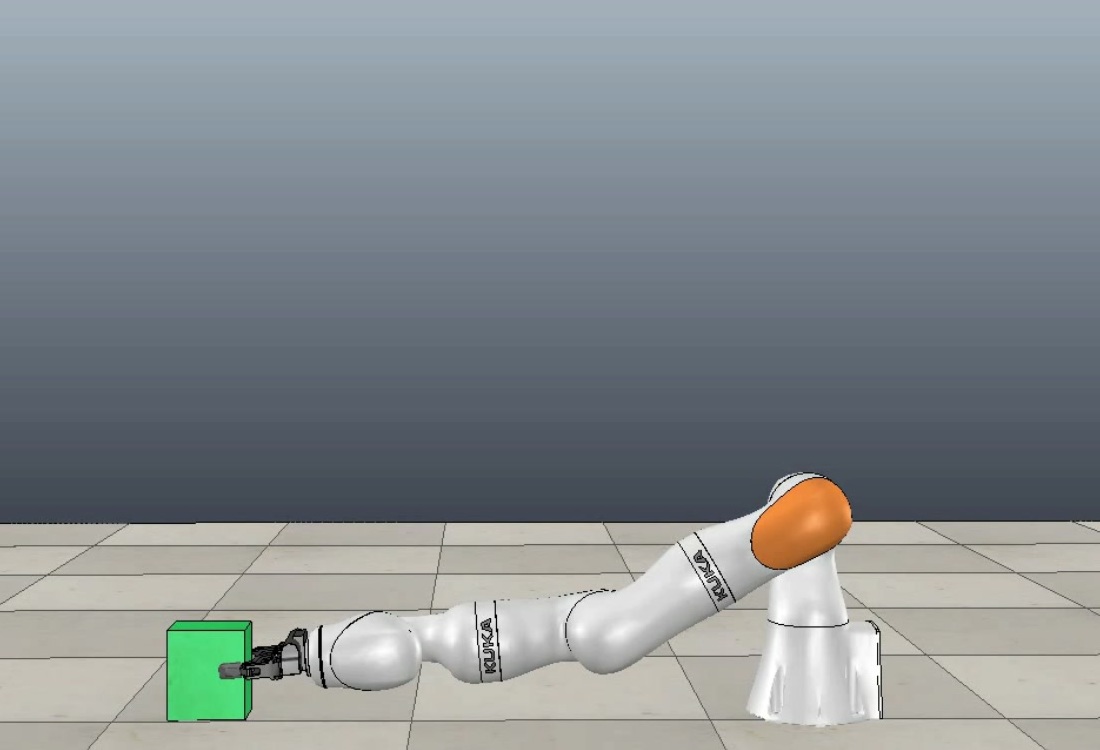}
		\caption{}
		\label{fig:kukaTorque0}
	\end{subfigure}
	\begin{subfigure}[b]{0.195\textwidth}
		\centering
		\includegraphics[scale=0.092]{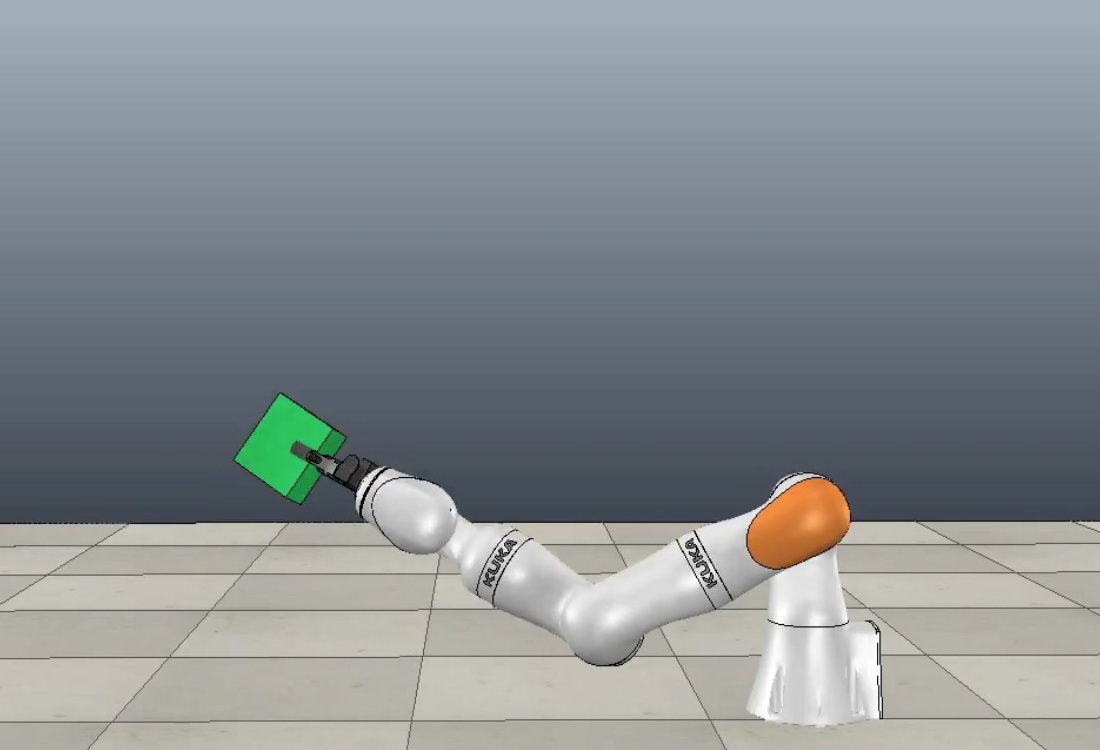}
		\caption{}
		\label{fig:kukaTorque1}
	\end{subfigure}
	\begin{subfigure}[b]{0.195\textwidth}
		\centering
		\includegraphics[scale=0.092]{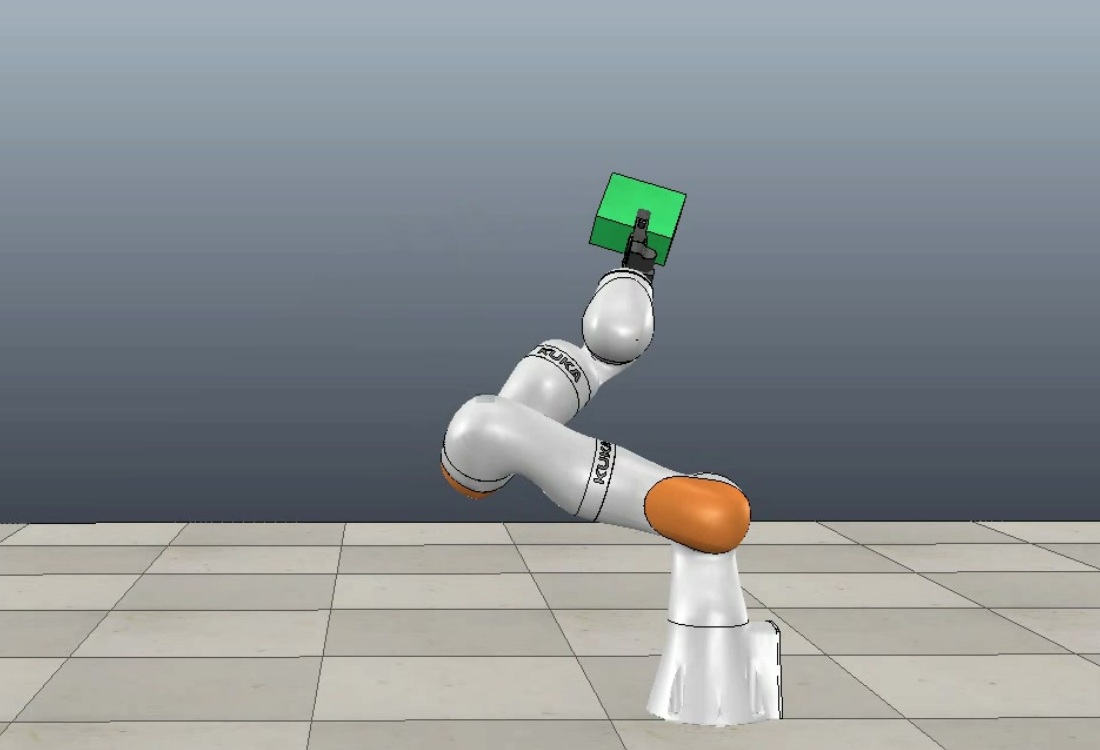}
		\caption{}
		\label{fig:kukaTorque2}
	\end{subfigure}
	\begin{subfigure}[b]{0.195\textwidth}
		\centering
		\includegraphics[scale=0.092]{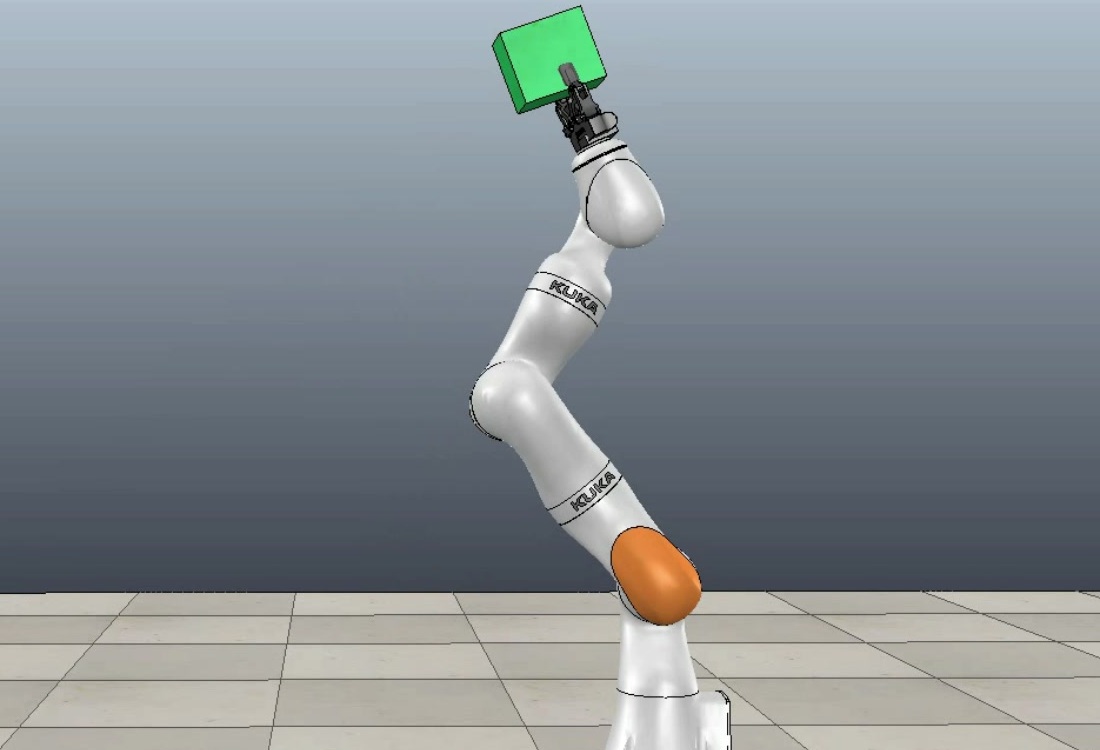}
		\caption{}
		\label{fig:kukaTorque3}
	\end{subfigure}
	\begin{subfigure}[b]{0.195\textwidth}
		\centering
		\includegraphics[scale=0.092]{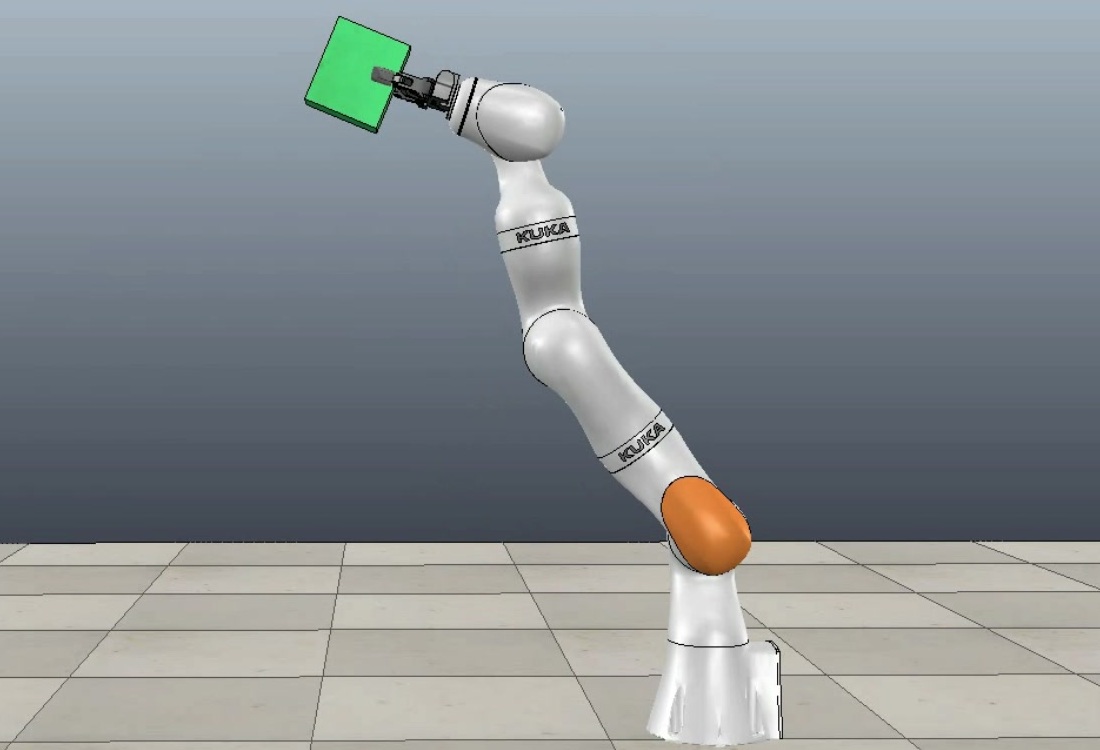}
		\caption{}
		\label{fig:kukaTorque4}
	\end{subfigure}
	\begin{subfigure}[b]{0.195\textwidth}
		\centering
		\includegraphics[scale=0.092]{figures/kukaTorque0.jpg}
		\caption{}
		\label{fig:rekukaTorque0}
	\end{subfigure}
	\begin{subfigure}[b]{0.195\textwidth}
		\centering
		\includegraphics[scale=0.092]{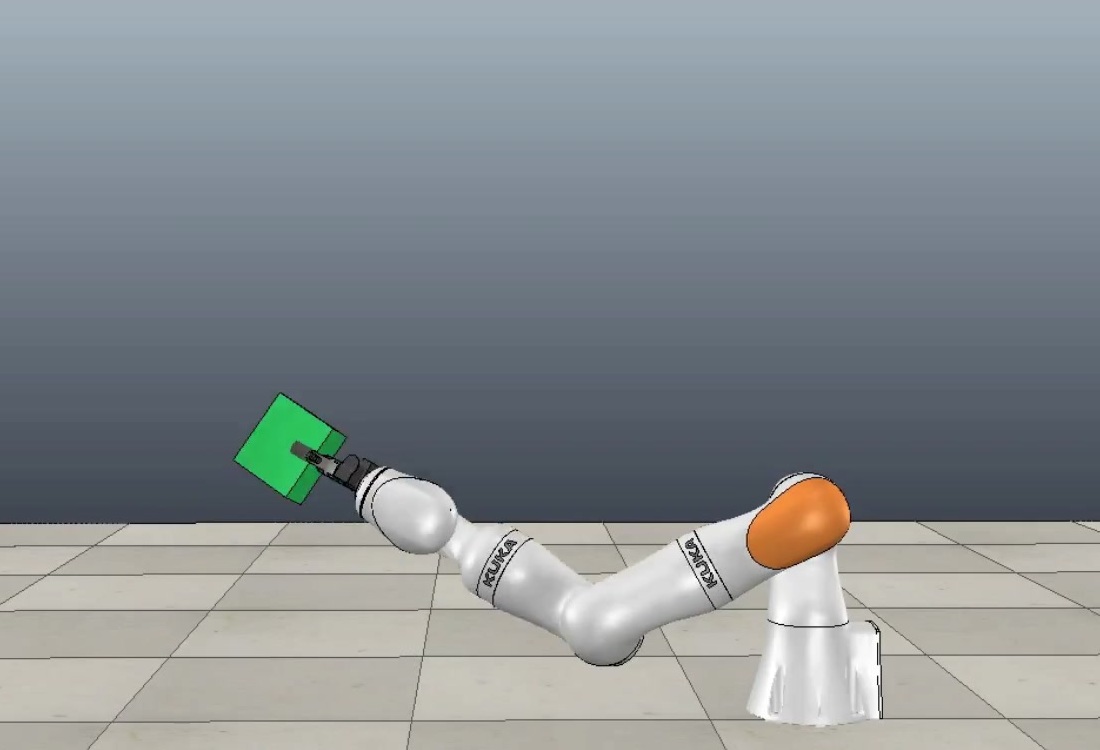}
		\caption{}
		\label{fig:rekukaTorque1}
	\end{subfigure}
	\begin{subfigure}[b]{0.195\textwidth}
		\centering
		\includegraphics[scale=0.092]{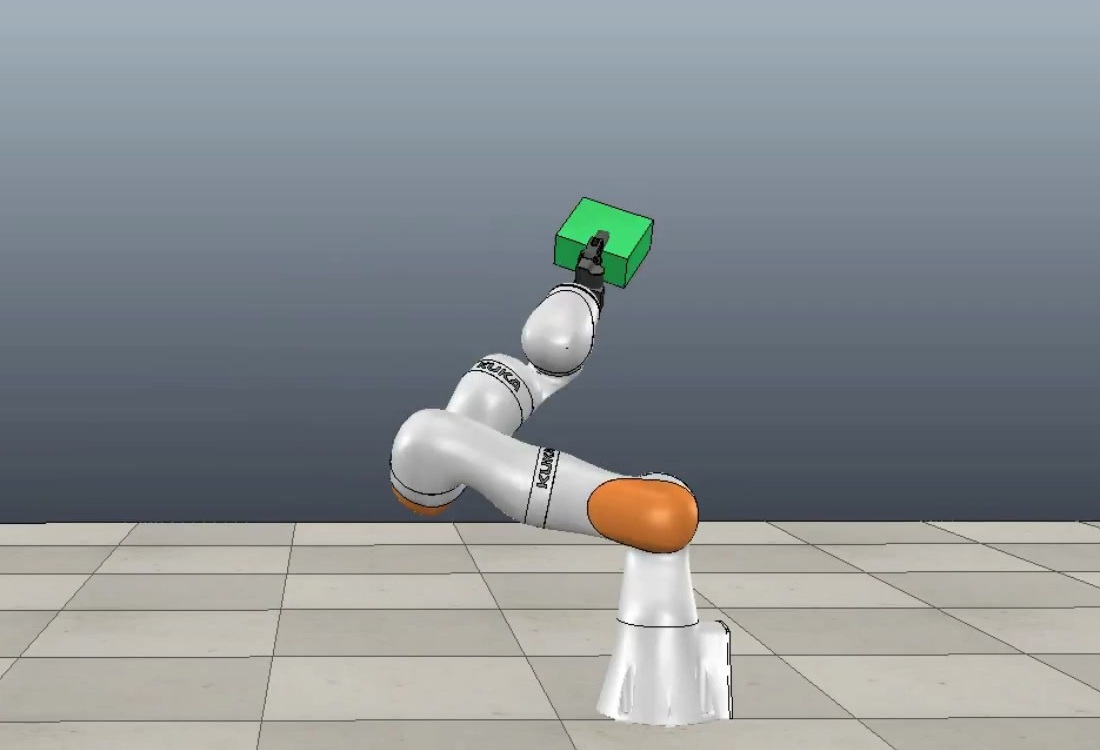}
		\caption{}
		\label{fig:rekukaTorque2}
	\end{subfigure}
	\begin{subfigure}[b]{0.195\textwidth}
		\centering
		\includegraphics[scale=0.092]{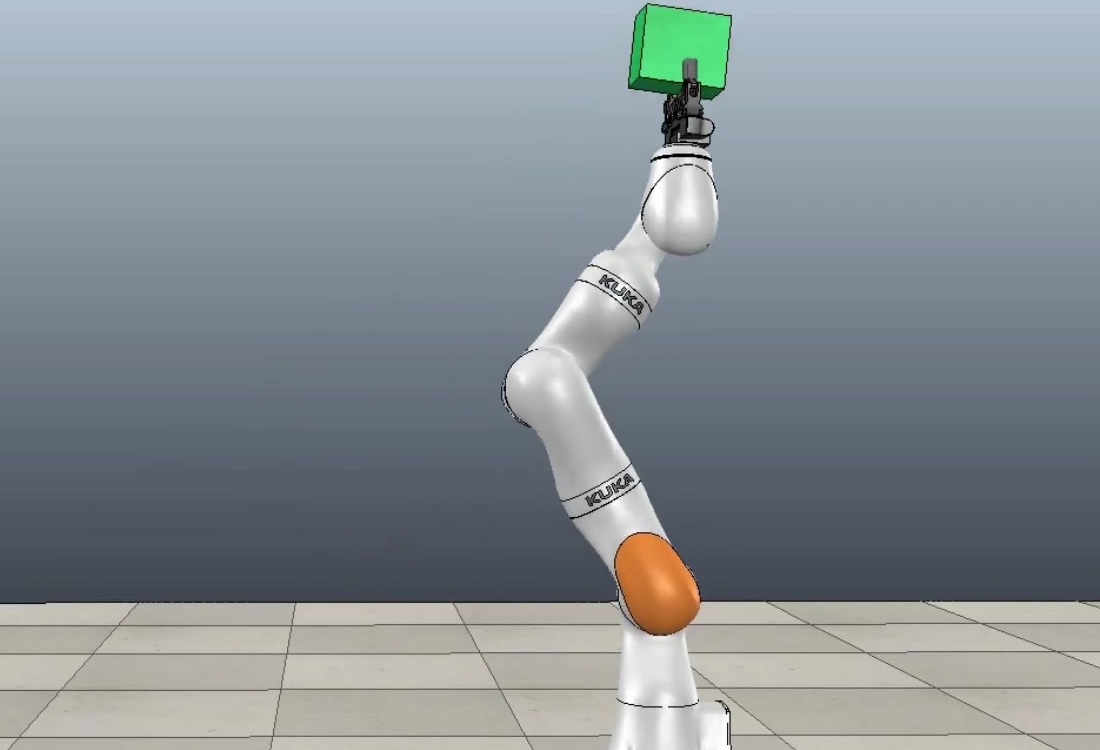}
		\caption{}
		\label{fig:rekukaTorque3}
	\end{subfigure}
	\begin{subfigure}[b]{0.195\textwidth}
		\centering
		\includegraphics[scale=0.092]{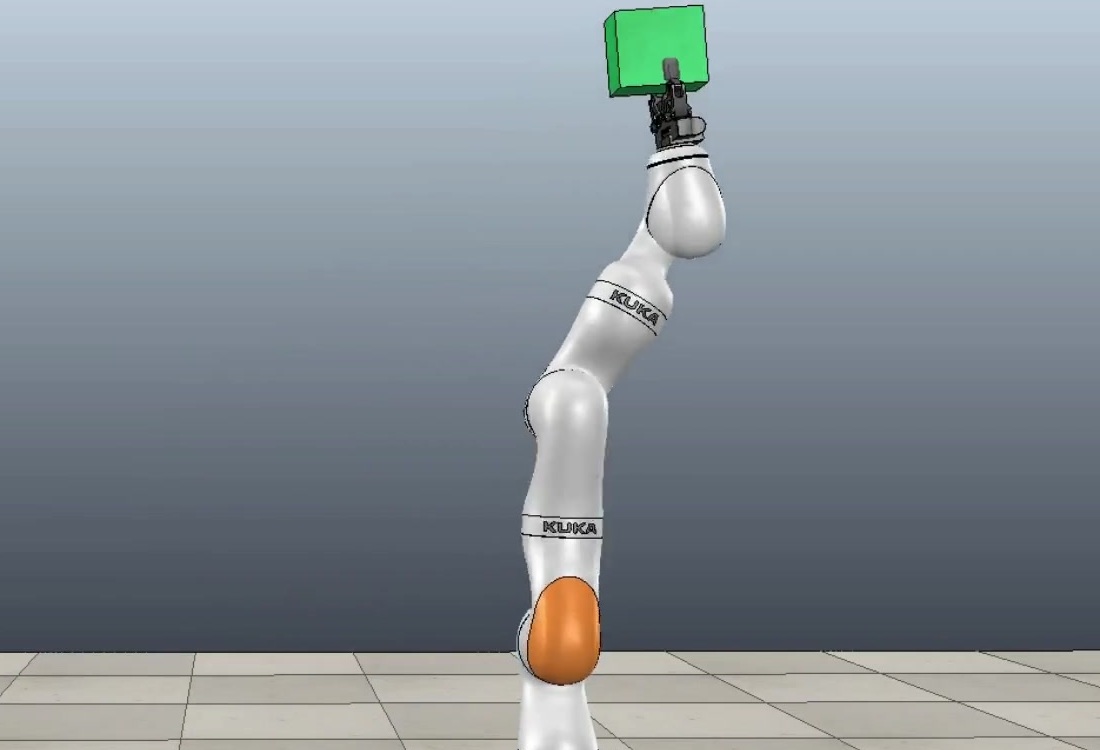}
		\caption{}
		\label{fig:rekukaTorque4}
	\end{subfigure}
	\caption{(a)-(e) show a solution obtained by the DFGP algorithm for kinodynamic motion planning task 2, where the objective is to find a trajectory for the KUKA robot to bring a block from the initial location as shown in (a) to the goal location as shown in (t). (f)-(j) show a solution obtained by the DFGP algorithm for the same task but with a minimum torque constraint applied. 
	(k)-(o) show a solution obtained by the iDFGP algorithm for the corresponding replanning problem with the new goal as shown in (o) while the minimum torque is required. }
	\label{fig:kuka-task2}
\end{figure*}

\begin{figure}[!htb]
	\centering
	\includegraphics[scale=0.4]{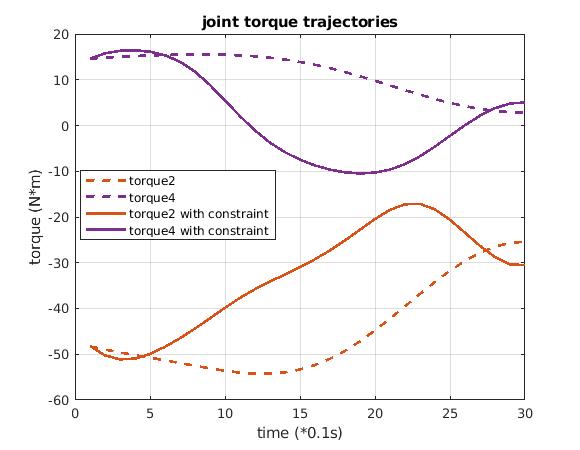}
	\caption{Torque trajectories for joint 2 (orange) and joint 4 (purple) of KUKA robot in task 2, where solid lines are results with minimum torque constraint, and dashed lines are results without minimum torque constraint.}
	\label{fig:torque}
\end{figure}

The KUKA LBR iiwa~\cite{kuka_webpage} is a lightweight industry robot with 7 actuated revolute joints. The tasks performed in our experiments are the following: 1) moving a block from one location to the desired goal location in the environment with obstacles, and 2) moving a block from one location to the desired goal location with minimum torque constraint.

We perform both kinodynamic motion planning tasks for the KUKA robot with the proposed DFGP algorithm, and the RRT-style methods Stable-Sparse-RRT(SST), Stable-Sparse-RRT*(SST*) and Asymptotically-Optimal-RRT(AO-RRT). For each task, we created 50 different tests with the same start configuration but randomly generated goal configurations. We ran 50 tests for each task with DFGP, SST, SST* and AO-RRT. However, due to the curse of dimensionality, none of the RRT-style methods were able to find a solution after more than 8 hours of computation. 

The benchmark results for task 1 and task 2 are shown in Table \ref{table:results_DFGP}. In task 1, we count one trial as a success if a collision free trajectory is found and all the dynamic constraints are satisfied. In task 2, one trial is considered successful if the goal is reached while all the dynamic constraints are satisfied and the total torque (total torque for all joints in each run) is less than the one without a minimum torque constraint. The average and maximum time to success is calculated with only the successful trials. With the minimum torque constraint, the average torque based on the successful trials is 1522.25 N*m, while it is 1747.91 N*m without this constraint. 

Figures~\ref{fig:kukaObs0} to~\ref{fig:kukaObs4} show a solution obtained by the proposed DFGP algorithm for task 1 with the goal configuration shown in Fig.~\ref{fig:kukaObs4}, from which we observe that the KUKA robot brings the block to the goal location while avoiding the desk which is initially above it.

Figures~\ref{fig:kukaTorque0} to~\ref{fig:kukaTorque4} show a solution obtained by the proposed DFGP algorithm for task 2 with the goal configuration shown in Fig.~\ref{fig:kukaTorque4}. We observe that the KUKA robot first moves towards the center to reduce the moment arm, then pushes upwards so that it can bring the block to the goal location with less torque applied as compared to the solution shown in Fig.~\ref{fig:kukaNoMinTorque0} to Fig.~\ref{fig:kukaNoMinTorque4}, where the start and goal configurations are the same, but with no torque constraint applied. 

The torque trajectories of joint 2 and joint 4 (the two joints that do most of the work in task 2) of the KUKA robot are plotted in Figure~\ref{fig:torque}, where the solid lines and dashed lines represent the torque trajectories with and without the minimum torque constraint, respectively. From the plots in Fig.~\ref{fig:torque} we observe that a more energy efficient motion plan is obtained by the DFGP algorithm with the minimum torque constraint.

\subsection{Fast Kinodynamic Motion Replanning for the KUKA}
\begin{table}[!htb]
	\caption{Results for Kinodynamic Motion Replanning.}
	\label{table:results_iDFGP}
	\begin{center}
		\begin{tabular}{| c | l | c | c | c |}
			\hline 
			\multicolumn{2}{|c|}{Kuka Tests} & DFGP-L & DFGP-W & iDFGP\\
			\hline
			\multirow{3}*{Task1}     
			& Success Rate (\%)& 86 & 92 & 80 \\
			& Average Time (s) & 1.98 & 1.12 & 0.119 \\
			& Maximum Time (s) & 3.33 & 3.09 & 0.200\\
			\hline 
			\multirow{3}*{Task2}     
			& Success Rate (\%) & 90 & 98 & 96 \\
			& Average Time (s) & 1.68 & 0.63 & 0.074\\
			& Maximum Time (s)   & 2.98 & 1.76 & 0.078\\
			\hline 
		\end{tabular}
	\end{center}
\end{table}

We change the goal configurations generated for each kinodynamic motion planning task to new goal configurations, and solve the replanning problem with both the batch version DFGP and the incremental version iDFGP. For the batch version DFGP, we ran the tests with two different initialization methods, one initialized with an acceleration-smooth straight-line (named DFGP-L), and the other uses a warm start with the result from the corresponding original planning problem (named DFGP-W). 

Replanning benchmark results for task 1 and task 2 are summarized in Table \ref{table:results_iDFGP}. The same rules are used to count the successful trials and calculate the average and maximum computation time as described in the original planning problem. 

From the results, we observe that in both task 1 and task 2, the iDFGP algorithm is about 8 to 10 times faster than the DFGP algorithm with the warm start initialization, and about 20 times faster than the DFGP algorithm with an acceleration smooth straight-line initialization, with only a small loss in the success rate in task 1. 

Our experimental setting for task 1 is similar to the one in~\cite{Zhao18icarcv_optControlMP}, and the results show that the efficiency of our DFGP algorithm is about the same as their method, and our iDFGP algorithm is about 20 times faster than either batch method. The failure of iDFGP algorithm in task 1 might be caused by the fact that iDFGP uses the original solution as initialization, which can be a bad choice if the goal has been changed dramatically. However, the DFGP algorithm using the warm start initialization did not suffer from any loss in success rate due to the fact that DFGP uses the Levenberg-Marquardt optimizer, which is a more powerful optimization algorithm compared to the one used by iDFGP, and is able to provide appropriate step damping to help converge to better results~\cite{Dellaert17fnt_fg}. The average torques calculated based on the results of the successful trials for DFGP-L, DFGP-W, and iDFGP were 1466.29 N*m, 1490.20 N*m, and 1489.03 N*m, respectively.

Figures~\ref{fig:rekukaObs0} to~\ref{fig:rekukaObs4} show a solution obtained by the proposed iDFGP algorithm for the replanning problem in task 1, where the goal is changed from Fig.~\ref{fig:kukaObs4} to Fig.~\ref{fig:rekukaObs4}. Since the environment is the same and the goal is not dramatically changed, the iDFGP algorithm is able to update the trajectory with an incremental Bayes tree solver, and find a solution which is very close to the original trajectory with only the part near the goal being modified. A similar effect can be observed in Fig.~\ref{fig:kuka-task2}, where Figures~\ref{fig:rekukaTorque0} to~\ref{fig:rekukaTorque4} show a solution by the proposed iDFGP algorithm for the change of goal replanning problem in task 2. 

\section{Conclusion}
We consider the kinodynamic motion planning as an optimal control problem and efficiently solve using the factor-graph-based GTSAM solver, obtaining performance on par with the state of the art for initial planning. We are able to perform fast replanning by taking advantage of the incremental Bayes tree solver, which avoids resolving the entire problem if only part of the trajectory needs to be updated. 

In the future work, we would like to explore more initialization strategies to improve the success rate of the proposed kinodynamic motion planning algorithm and help the optimization solver to converge to a better local optimal solution.

\clearpage
\bibliographystyle{plainnat}
\bibliography{references.bib}

\end{document}